\documentclass[lettersize,journal]{IEEEtran}
\usepackage{amsmath,amsfonts}
\usepackage{amsmath}
\usepackage{array}
\usepackage[caption=false,font=normalsize,labelfont=sf,textfont=sf]{subfig}
\usepackage{textcomp}
\usepackage{stfloats}
\usepackage{url}
\usepackage{verbatim}
\usepackage{graphicx}
\usepackage{cite}
\usepackage{tikz}
\usepackage{multicol}
\usepackage{booktabs}
\usepackage{multirow}
\usepackage{bbm}
\usepackage{physics}
\usepackage{algpseudocode}
\usepackage{algorithm}
\usepackage{hyperref}
\usepackage[normalem]{ulem}
\usepackage{multicol}

\algnewcommand\algorithmicforeach{\textbf{for each}}
\algdef{S}[FOR]{ForEach}[1]{\algorithmicforeach\ #1\ \algorithmicdo}
\usetikzlibrary{arrows}
\usepackage[
separate-uncertainty = true,
multi-part-units = repeat
]{siunitx}
\begin{document}
	
	\title{ReMAV: Reward Modeling of Autonomous Vehicles for Finding Likely Failure Events }
	\author{Aizaz Sharif and Dusica Marijan}
	
	
	
	\maketitle
	
	\begin{abstract}
Autonomous vehicles are advanced driving systems that are well known to be vulnerable to various adversarial attacks, compromising vehicle safety and posing a risk to other road users. Rather than actively training complex adversaries by interacting with the environment, there is a need to first intelligently find and reduce the search space to only those states where autonomous vehicles are found to be less confident. In this paper, we propose a black-box testing framework \textit{ReMAV} that uses offline trajectories first to analyze the existing behavior of autonomous vehicles and determine appropriate thresholds to find the probability of failure events. To this end, we introduce a three-step methodology which i) uses offline state action pairs of any autonomous vehicle under test, ii) builds an abstract behavior representation using our designed reward modeling technique to analyze states with uncertain driving decisions, and iii) uses a disturbance model for minimal perturbation attacks where the driving decisions are less confident. Our reward modeling technique helps in creating a behavior representation that allows us to highlight regions of likely uncertain behavior even when the standard autonomous vehicle performs well. This approach allows for more efficient testing without the need for computational and inefficient active adversarial learning techniques. We perform our experiments in a high-fidelity urban driving environment using three different driving scenarios containing single- and multi-agent interactions. Our experiment shows an increase in 35\%, 23\%, 48\%, and 50\% in the occurrences of vehicle collision, road object collision, pedestrian collision, and offroad steering events, respectively by the autonomous vehicle under test, demonstrating a significant increase in failure events. We compare ReMAV with two baselines and show that ReMAV demonstrates significantly better effectiveness in generating failure events compared to the baselines in all evaluation metrics. We also perform an additional analysis with prior testing frameworks and show that they underperform in terms of training-testing efficiency, finding total infractions, and simulation steps to identify the first failure compared to our approach. The results show that the proposed framework can be used to understand the existing weaknesses of the autonomous vehicles under test to only attack those regions, starting with the simplistic perturbation models. We demonstrate that our approach is capable of finding failure events in terms of increased collision and offroad steering errors compared to standard AV driving behavior.

	\end{abstract}
	
	
	\section{Introduction}

	Autonomous Vehicles (AVs), also known as self-driving cars, are complex technologies that are susceptible to failure~\cite{Garcia}. The development of deep learning-based AVs is a challenging problem~\cite{Riccio, MLTest} due to the numerous available architectures. Software for AVs is based on artificial intelligence (AI) models, which have seen significant progress in recent years~\cite{R5,R23}. However, the validation of autonomous systems remains a challenging area despite the advances made. 
	
	One of the main techniques for validating AVs is to use adversarial noise attacks~\cite{R4},~\cite{R8},~\cite{Generating},~\cite{R11},~\cite{R44} to test the robustness and reliability of AV software. These attacks involve malicious actors adding noise or distortion to the inputs of AV models. In the context of AVs, they could cause the vehicle's object detection and control systems to misinterpret the environment, leading to potentially dangerous situations. The application of these attacks helps to identify vulnerabilities and weak points in AV systems, thus improving their safety and performance. By mimicking real-world scenarios, adversarial attacks allow us to evaluate how software might behave under various unpredictable conditions, contributing to the overall objective of software engineering to build efficient, reliable, and secure systems.

	While testing single-agent AVs is challenging, the problem is even more complex in the presence of other agents~\cite{R13}. AV testing in a multi-agent setting is a complex problem since it is hard to find which states of interactions will most likely lead AVs to fail. In simulation-based testing and validation of AVs, we usually perform many simulation scenarios, and if the AVs are trained in the same environment, they are likely to drive well. The idea of adversarial testing for exposing faults in AVs works because, in a collaborative as well as the competitive driving world, not all agents are perfect, so we need adversarial agents within multi-agent interactions. For adversarial testing, there are also several works in the field of creating adversary-driving and non-playable character (NPC) agents within the same environments~\cite{R8,R4}. Such adversarial agent behaviors are also missing while training AVs with online/offline datasets~\cite{codevilla2018offline,codevilla2019exploring}.
	
	For training adversarial agents with the goal of producing possible failure scenarios, they are trained in a black-box 2-player zero-sum game~\cite{R6} or transformed into existing NPC agents around the AV as aggressive collaborative / competitive adversaries~\cite{R8,R28,9636072}. Such adversaries are usually trained against a known AV in a two-player game with a hand-made reward function~\cite{R34} and hence are very restrictive in approach. Therefore, this method cannot be scaled and a lot of training computation is required to make the adversary efficient. In case we retrain our AVs using the missing adversarial examples~\cite{10043282}, the adversary must be modified as a zero-sum game. In a multi-agent scenario, deciding how many adversaries, if adversaries are also driving and NPC agents to train, and which type of adversary to train are some of the critical decisions to be made. Without first identifying the existing weaknesses of the AV under test, this approach is highly computationally inefficient and time consuming. Another challenge in testing AVs using adversaries is that the search space for AV systems is high-dimensional and stochastic~\cite{Alessio}, and it is not possible to cover all possible states. Therefore, we need to narrow down the search space to only those interactions in which autonomous vehicles are less confident in taking the correct control actions. The question arises: Can we simply utilize the existing simulation data of AVs and use those trajectories to find those likely failure events?

	All the limitations mentioned above lack \textit{behavior representation} of a state-action-only distribution. This helps us to understand and reduce the search space for the most likely failure states to plan minimal required attacks~\cite{lin2017tactics}. There is a need for a scalable black-box testing approach to test any AV whether using input perturbation or transforming NPCs into adversaries on the spot. Once we identify the uncertain behavior of the AV, we can prove that the existing simplistic methods of noise disturbance as adversarial attacks might just be a good start for validating the safety of AVs instead of training complex adversarial models.

	To address these challenges, in this paper, we propose a framework \emph{Reward Modeling of Autonomous Vehicles (ReMAV) for Finding Likely Failure Events} which is a novel approach for testing AVs. We design this framework as a black-box approach where any sort of autonomous driving agent's offline data can be used for testing AVs. In our case, we are designing this framework as a black-box method that only deals with state action pairs, either collected recently or stored in the past through normal simulation testing. Regardless of whether AVs are trained on reinforcement learning (RL) or other deep learning based algorithms, we treat them as a black-box, and only consider their action distributions and decision-making. This provides scalability and flexibility to any kind of testing simulation environment. Once we have offline trajectories, we utilize inverse reinforcement learning (IRL) to obtain a reward model for the autonomous vehicles under test. IRL has been applied mainly for transfer learning of expert policies, but has not yet been used for behavior representation in testing autonomous vehicles.
	
	In our work, we start by using the baseline IRL algorithm and modifying it to represent the behavior of AVs as a reward modeling technique in three driving scenarios for evaluation. Our novel contribution is the use of IRL in behavior representation for testing and validating autonomous vehicles, which to the best of our knowledge has not been performed before. Rather than actively training adversaries by interacting with the environment, our approach uses \textit{offline trajectories} collected within the standard simulation to first analyze the state of the AV and then determine appropriate \textit{thresholds} for finding the probability of failure events. This approach allows for more efficient testing without the need for active adversarial training. We can use the reward model to quantify the behavior of AVs in various driving scenarios to assess their robustness. By analyzing this model, we can understand driving behavior and identify weak driving events to confirm their resilience.


	The key research contributions in this paper are as follows:
	
	\begin{itemize}
		\item We introduce a novel testing framework to model abstract unsupervised reward functions by utilizing the collected offline trajectories of AVs in a scenario.
		\item We design a reward modeling technique using inverse reinforcement learning to find the probability of failure events.
		\item We experimentally show that reward modeling allows one to understand the behavior of AVs under test by analyzing their behavior distribution.
		\item We demonstrate that using a disturbance model for minimal perturbation attacks on input sensor data and existing NPCs creates challenging adversarial events for the same AVs for testing.
		\item We perform an extensive evaluation to show that the ReMAV methodology outperforms the baselines in all evaluation metrics in three driving scenarios.
		\item We also compare the failure detection efficiency of our framework with existing relevant frameworks.

	\end{itemize}

	\section{Problem Setup}
	
	In our work, we design a black-box testing framework to test the robustness of the AVs under test. We do so by modeling their behavior representation to first identify states that might lead to likely failure events. Once identified, we use the disturbance model to add minimal perturbations. In this section, we define the scope and set assumptions before explaining the methodology of our framework.

	\subsection{Failure Definition}
	In a black-box safety validation approach, our task is to identify the subset of trajectories that are most likely to cause the AV under test to experience failure events. Assuming $x \in \mathbb{R}$ defines a disturbance input (such as perturbation attacks on AV input sensor data or existing NPC actions) to a sequential black-box system under test and $y \in \{0,1\}$ is a boolean output of the failure event, we can formally define the estimation of failure probability as:

	$$ P_{\text{failure}} = \mathbb{E}_{p(x)} \mathbbm{1}(y \ne 0). $$

	In a multi-agent environment where cars and pedestrians interact, failure can be defined as the occurrence of collisions and offroad steering errors. A collision occurs when the path of a car intersects the path of a pedestrian and the car and pedestrian come into contact. A collision within vehicles can be defined as the occurrence of two or more agents occupying the same physical space at the same time. The same can be said for collisions made with any road object. Offroad steering errors occur when a car deviates from its intended path and goes offroad, potentially endangering pedestrians and other driving agents. To determine failure in this multi-agent environment, we can monitor collision and offroad steering events in our scope of work and consider them as failure states.

	\subsection{AV under test}
	We also need to specify the properties of AV under test in order to define the scope of our work.
	
	\subsubsection{Markov Decision Processes (MDPs) based Deep Reinforcement Learning (DRL) approach}
	
	Our testing framework is designed to evaluate the performance of AVs using offline data. We consider AVs as a black-box model, and our framework only requires access to the input state observations and output actions of AV models stored as offline trajectories. We used an existing multi-agent urban driving framework to facilitate training DRL-based AV policies. This allowed us to train standard AV driving agents with an MDP-based DRL approach.
	
	\subsubsection{Vision-based driving models}
	Our framework trains and tests AVs as a vision-based driving model in urban driving scenarios. Each AV in a simulation receives 3D input sensor information and using an end-to-end approach, the models output an action that is passed to the simulator. Every agent's action returns a reward and a next input state that is based on the same AV models as a sequential MDP approach.
	
	\subsubsection{Non communicating model-free RL agents}
	AV policies based on DRL are trained using a model-free assumption and a Partially Observable Markov Decision Process (POMDP) approach. In a multi-agent driving environment, AVs are considered independent and competitive players who do not communicate with one another. This implies that they do not have access to each other's input state or shared information on weight parameters.
	
	\subsection{Reward Modeling Definition}
	Let $\tau_i = (s,a)$ be the trajectories of pairs of state actions of the AV under test. We define a reward model $R_{\psi}$ that uses offline trajectories and creates a function that represents the behavior of the AV under test. The objective of $R_{\psi}$ can thus be defined as
	
	$$ R_{\psi} = \max_{R} \sum_{i=1}^N \log p(\tau_i | R),$$

	where $p(\tau_i | R)$ is the probability of trajectory $\tau_i$ under the actual reward function $R$. Our goal is to use the learned reward model to obtain the AV behavior representation to identify thresholds $\beta$. The threshold hence can be used in simulation testing with the help of minimal perturbation attacks in order to find the most likely failures.
	
	\begin{figure*}[!htbp]
		\centering
		
		\includegraphics[width = 1.0\textwidth,height=0.24\textheight]{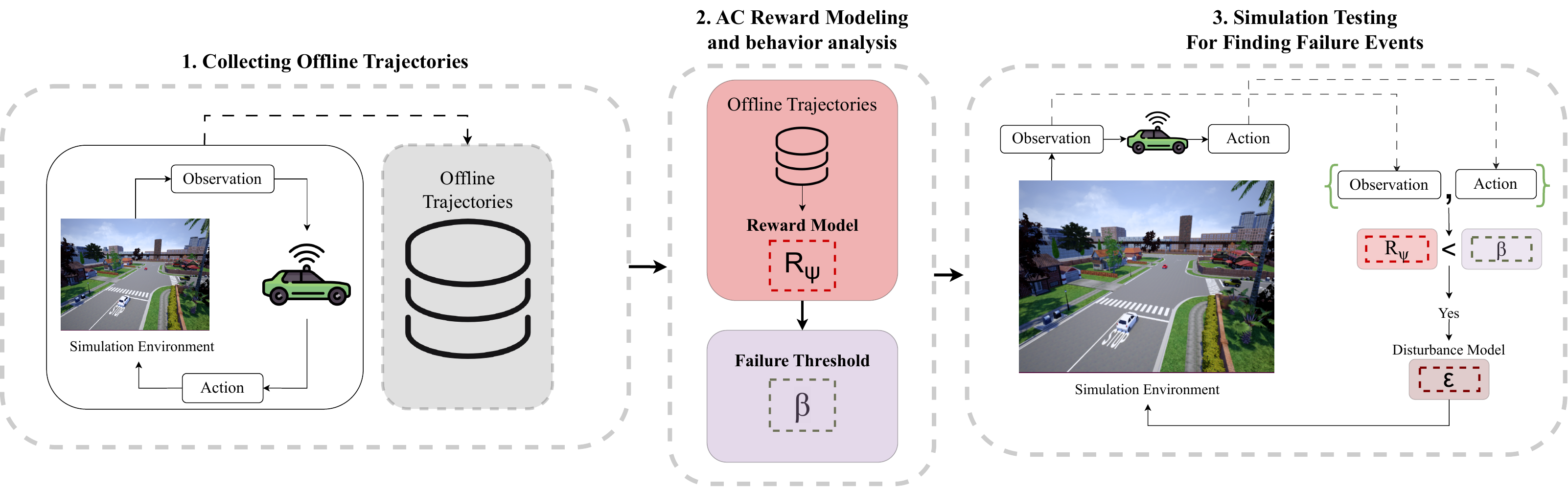}
		\caption{Illustration of ReMAV framework architecture for testing the robustness of AV driving policies in a multi-agent environment. The framework is divided into three steps: The left represents the first step where we use the AV under test to obtain its offline trajectories in different driving scenarios. The middle represents the phase where offline trajectories are used to obtain a behavior representation with the help of a reward model $R_{\psi}$. The same reward model is used to collect \{state, action, reward\} pairs to collect the required thresholds $\beta$ for testing. Last, the right side of the architecture shows the AV testing phase, where noise perturbations are added only to those state-action interactions where AV feels uncertain even when their driving behavior seems normal.}
		\label{fig:ReMAVArchitecture}
	\end{figure*}
	
	\section{ReMAV Framework}\label{sec:ReMAV}

	In the following section, we utilize these concepts to present the proposed framework named ReMAV and the methodology for testing AVs using reward modeling and behavior representation. An overview of the three-step methodology using the ReMAV framework is illustrated in Figure~\ref{fig:ReMAVArchitecture}.
	
	In ReMAV, we consider DRL-based AV under test in a vision-based urban driving environment. As a first step of our framework, we use existing AVs to collect offline state-action pairs to record multiple trajectories using normal simulation. In the second step, we train the inverse reward modeling approach to learn an abstract unsupervised reward function. The learned reward model is used to record AV behavior in order to go through statistical analysis to find thresholds beyond where AV might go into a failure state. In the last step of the framework, we perform simulation testing using the threshold per scenario and add minimal perturbation attacks in order to see whether AV leads to a failure event or not.

	\subsection{Step 1: Collecting Trajectories of AVs}
	
	Our framework can be used to test any AV model, since we have designed it in a black-box approach. Our methodology starts with collecting offline trajectories of the AV policy $\pi_{AC}$ under test, which is pre-trained in an online or offline simulation setting using any AI-based approach. For this work, we also train a vision-based AV in a competitive multi-agent environment before collecting its offline state-action pairs.

	\subsubsection{AV Architecture details}
	\paragraph{PPO for AV driving policy}\label{sec:PPOAV}

	Our standard AV agent utilizes the PPO algorithm~\cite{R55} as a policy gradient method to acquire a driving policy, $\pi_{AC}$, through interactions with a simulated environment during each training episode. This approach enables on-policy learning within the simulation, rather than relying on a dataset (replay buffer) for learning. PPO also offers the ability to update the policy in a stable manner during the learning process, even when faced with changes in data distributions. Furthermore, PPO addresses the challenge of initializing a large hyperparameter space during the learning process.

	\paragraph{Deep Neural Network Model}\label{sec:mnih15}
	
\begin{figure}[!t]
		\centering
		
		\includegraphics[width = 0.48\textwidth,height=0.2\textheight]{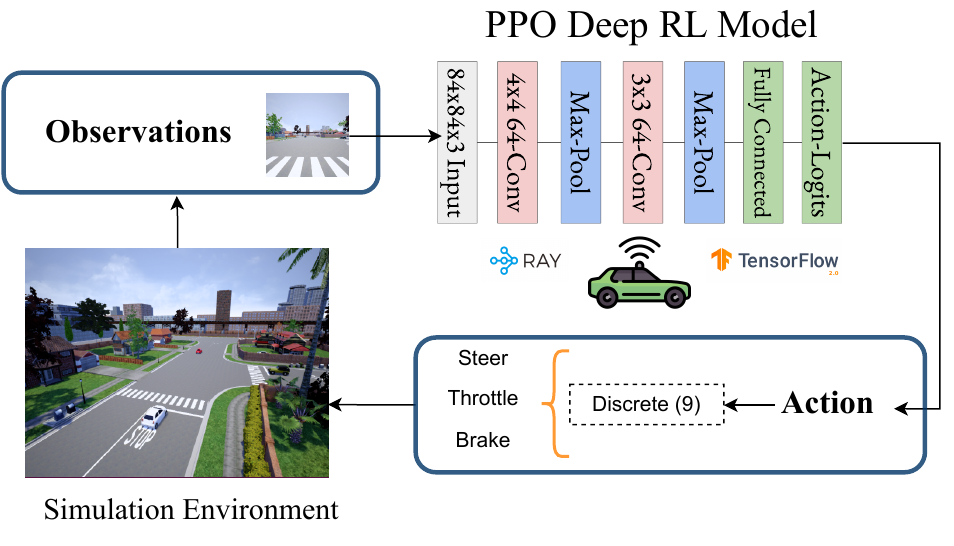}
		\caption{End-to-end AV Driving using DRL for AV agents. AV receives an input image of 168$\times$168$\times$3 which is passed to the PPO based DRL model. The actions are selected in the output layer of the AV agent and performed in the next time step of the simulation to obtain a reward and a new observation state.}
		\label{fig:deepRL}
	\end{figure}

	A visual description of the DRL architecture for AV, including the input, hidden, and output layers, is shown in Figure~\ref{fig:deepRL}. The DRL architecture consists of the input layer that receives a partial observation of 168$\times$168$\times$3 pixel images captured by cameras mounted in front of the driving model. These images are processed through hidden and convolutional layers before reaching the output layer, which generates control commands. The AV driving agent has nine discrete values available at the output layer to make control decisions. These values can be grouped into three main actions named Steer, Throttle, and Brake, which are passed to the driving simulator.

	\paragraph{Reward Function for AV}\label{sec:Reward}
	
	The AV agent follows the MDP formulation and therefore, at each time step, the driving models collect trajectories of $(S, R, A)$. $R$ is the reward gained in return for the actions chosen by the driving car policy function, given the input observations.

	 AV policies are trained with the goal of reaching the desired destination safely as close as possible. The AV agent's reward can be described as:
	$$R_{\text{AV}} = \alpha_1\Delta (D_t) + \alpha_2(Y_t)-\alpha_3(CV_t + CO_t) -\alpha_4(OS_t),$$
	
	where $\Delta D_t$ is the distance covered, $Y_t$ is the forward speed of the agent, $CV_t$ and $CO_t$ are the Boolean values that indicate whether there is any collision with other vehicles and objects in the environment, and $OS_t$ refers to driving offroad represented as boolean. $\alpha$ are the coefficients with $\alpha_1$=1, $\alpha_2$=1, $\alpha_3$=100, and $\alpha_4$=0.5. The coefficient values are selected judiciously in order to train the AV agent. The equation above shows that AV driving policies are greatly affected by steering errors and collisions that occur off the road.
	
	\subsubsection{Offline Trajectories}
	
	When it comes to testing AVs in our methodology, the first step is to collect state-action trajectories. These trajectories are essentially a record of the car's behavior in different scenarios, and they are crucial for training and testing machine learning algorithms that power the AV's decision-making processes.
	
	We collect state-action trajectories by running normal simulation episodes. These simulations involve running the car through three scenarios in a controlled environment. By carefully controlling the conditions of these scenarios, we collect data on how the car behaves in different situations, such as when it encounters obstacles, changes lanes, or interacts with other vehicles on the road. To collect state-action trajectories, we must record both the AV's state (such as its vision-based observational state per timestep) and the actions it takes (such as accelerating, braking, or turning).

	In DRL, a trajectory is a sequence of state-action pairs that an agent encounters as it interacts with the environment. Let us define a trajectory as $ \tau = [(s_1, a_1), (s_2, a_2), ..., (s_T, a_T)] $, where $s_t$ is the state in step $t$ and $a_t$ is the action taken by the AV policy at time-step $t$, and $T$ is the final step of the trajectory. To collect pairs of state-action trajectories, we use the AV policy $\pi_{AV}$, which is a mapping from states to actions. At each timestep, $t$, the AV agent selects an action $a_t$ according to the policy $\pi_{AV}$, and the environment transitions to a new state $s_{t+1}$.
	
	The process of collecting state-action pairs of trajectories can be defined as follows:
	
	\begin{enumerate}
		\item Initialize an empty set of trajectories $D = \{\}$.
		\item For each episode:
		\begin{enumerate}
			\item Initialize an empty trajectory $\tau = [(s_1, a_1)]$.
			\item Observe the initial state $s_1$.
			\item Select an action $a_t$ according to the policy $\pi_{AV}$.
			\item Observe the next state $s_{t+1}$ and the reward $r_t \in R_{\text{AV}}$.
			\item Append the state-action pair $(s_t, a_t)$ to the trajectory $\tau$.
			\item Append the trajectory $\tau$ to the set of trajectories $D$.

		\end{enumerate}
		\item Repeat for each time-step $t$ until the end of the episode:
		\item Return the set of trajectories $D$.
	\end{enumerate}

	Collecting state-action pairs of trajectories enables our next step in the methodology where an IRL-based agent learns from its interactions with the environment for the behavior representation of AV.

	\subsection{Step 2: Reward modeling for AV behavior representation}
	
	Once we have offline trajectories of the AV under test, we perform the second step in the framework where we learn a reward model using the inverse reinforcement learning technique. Learning a reward model will help us obtain the behavior representation of the same AV in order to analyze its driving behavior in different scenarios.
	
	Here we first give a brief overview of the adversarial inverse reinforcement learning (AIRL) algorithm~\cite{fu2017learning}. Afterwards, we present the technical choices of the proposed testing methodology. AIRL is an inverse reinforcement learning algorithm that aims to learn a reward function from a set of expert trajectories. The learned reward function can then be used to train a policy that mimics the behavior of the expert.

	\subsubsection{AIRL in our methodology}
	
	Since our goal is to validate the safety and robustness of AV, we formulate this problem by considering AV as non-experts.
	Despite many advancements made in the AIRL research, we find the baseline AIRL algorithm suitable for our framework since we aim to learn a reward model out of a single non-expert driving agent that is part of a multi-agent environment. We also modify AIRL for training reward function in a multi-agent based complex urban driving environment. Additionally, we also implement AIRL specifically for our case in order to handle high-dimensional vision-based tensors by adding a feature extraction layer. We use the reward function to obtain an abstract distribution of the driving behavior of AV in a particular environment for further analysis. We also use this to quantify the behavior of AV to compare AV decision-making in different situations in the third step of our framework methodology.
	
	To formalize the problem, let us consider an MDP with a state space $S$, an action space $A$, and a discount factor $\gamma$. The objective is to learn a policy $\pi_G$ that maximizes the expected sum of discounted rewards:
	
	$$\mathbb{E}*{\tau \sim \pi_{G}}[R(\tau)] = \mathbb{E}*{\tau \sim \pi_{G}}[\sum_{t=0}^{\infty}\gamma^t r(s_t, a_t)]$$
	
	where $\tau = \{(s_t, a_t)\}$ is a trajectory, $r(s_t, a_t)$ is the reward function and $\mathbb{E}*{\tau \sim \pi_{G}}$ denotes the expectation taken over the trajectory generated by the policy $\pi_{G}$. Let us denote the true reward function by $R(\tau)$, and the learned reward function as $R_{\psi}$. The goal of AIRL is to learn the parameters $\theta$ of $R_{\psi}$ such that the policy that maximizes the expected sum of discounted rewards under $R_{\psi}$ is similar to the expert policy.

	The optimization problem of AIRL can be formulated as a two-player minimax game where the first player is the policy $\pi_{G}$ and the second player is a discriminator $D_\psi$ that tries to distinguish between expert trajectories and generated trajectories. The objective of the policy is to generate trajectories that are indistinguishable from expert trajectories, while the objective of the discriminator is to correctly classify the trajectories. The discriminator is trained to maximize this objective, while the policy $\pi_{G}$ is trained to minimize it. The discriminator in the AIRL algorithm is a binary classifier that distinguishes between expert and novice trajectories. Specifically, it is a function of the state-action pairs, denoted by $D_{\psi}(s,a)$. The discriminator is trained to maximize the probability of correctly classifying the expert and novice trajectories. This is accomplished by minimizing the binary cross-entropy loss function:
	\begin{multline*}
	L_{D_\psi} = \sum_{t=0}^T -\mathbb{E}_{\tau \sim \pi_{AV}}[\text{log}(D_{\psi}(s,a))] \\ - E_{\tau \sim \pi_G}[\text{log}(1 - D_{\psi}(s,a))],
	\end{multline*}

	where $L_{D_\psi}$ gives a loss between expert demonstrations from $\pi_{AV}$ and generated samples from $\pi_{G}$.

	The reward $R_\psi$ is therefore calculated using the following function:
	
	\begin{equation}
	R_\psi = \text{log}(D_{\psi}(s,a) - \text{log}(1-D_{\psi}(s,a))]
	\end{equation}
	
	
	Finally, the optimal policy, $\pi_{G}*$, is obtained by maximizing the expected reward value $R_\psi$. 
	
	$$\pi^{*}_{G} = \mathbb{E}_{\pi_{G}}\sum_{t=0}^T[R_\psi].$$
	
	Thus, the discriminator in AIRL is optimized to distinguish between expert and novice trajectories and is used to learn a reward model that guides the agent toward generating expert-like behavior. This is accomplished without the need for explicit reward values. The algorithm for AIRL to learn $R_\psi$ in our framework can be presented in Algorithm~\ref{alg:1}.

	\begin{algorithm} \caption{Reward Modeling from AV Offline Trajectories}\label{alg:1}
		\begin{algorithmic}[1]
			\Require Reward Model $R_\psi$.
			\State Obtain expert trajectories $\tau_i^{AC}$
			\State Initialize policy $\pi_G$ and discriminator $D_\psi$.
			\ForEach {$t \in \{1, \ldots, \mathrm{N}\} $} 
			\State $\quad$ Collect trajectories $\tau_i=\left(s_0, a_0, \ldots, s_T, a_T\right)$ by executing $\pi_{G}$. 
			\State $\quad$ Train $D_{\psi}$ via binary logistic regression to classify expert data $\tau_i^{AC}$ from samples $\tau_i$. 
			\State Update reward $R_{\psi}\left(s, a, s^{\prime}\right) \leftarrow \log D_{\psi}\left(s, a\right)-\log \left(1-D_{\psi}\left(s, a\right)\right)$ 
			\State Update $\pi_{G}$ with respect to $R_{\psi}$.
			\EndFor
		\end{algorithmic}
	\end{algorithm}

	\paragraph{AIRL Architecture Design}

\begin{figure}[t]
		\centering
		
		\includegraphics[width = 0.48\textwidth,height=0.2\textheight]{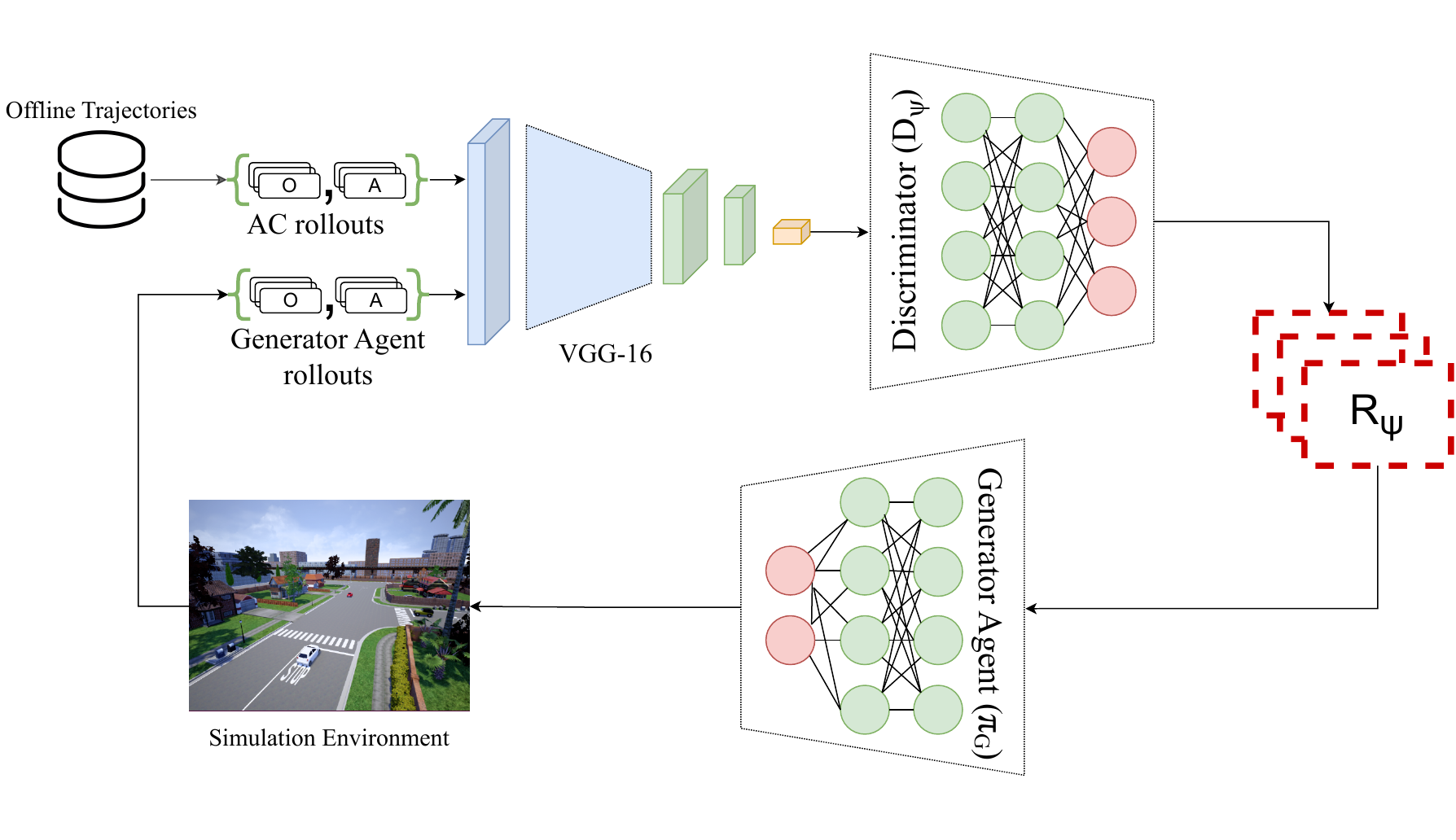}
		\caption{Illustration of the AIRL design as part of the ReMAV framework. Both AV offline trajectories and Generator rollouts are first passed through a VGG-16 layer in order to reduce dimensions to only important features. The Discriminator tries to classify between real (non-expert AV) and fake state-action pairs. The prediction is then transformed into a reward value $R_\psi$ that is also used to calculate the advantage function to improve the generator agent policy. The generator agent finally trains on a few samples to take better actions in the simulation environment in the next loop. }
		\label{fig:IRL}
	\end{figure}

	An overview of the reward modeling process can be illustrated in Figure~\ref{fig:IRL}. The offline trajectories collected in the first step of the framework are sampled with the dimension 168 $\times$ 168 $\times$ 3 and passed through the VGG-16 model~\cite{simonyan2014very} to reduce the dimensions of the input to 12800 as well as to obtain high-quality feature space. At the same time, the PPO-based generator policy $\pi_{G}$ takes some action against the observations of the simulation, and these state-action pairs are also passed through a VGG-16 layer to obtain a feature space of 12800. Both AV and $\pi_{G}$ based state-action pairs are passed in batches through a discriminator neural network which tries to classify AV data as real and $\pi_{G}$ based actions as false. The prediction from the discriminator is converted into a reward value as $R_\psi$. $R_\psi$ also recovers the advantage $A(s, a)$, which is used in the loss to update the generator agent. After updating the $\pi_{G}$ policy, sampling is performed in order to train the generator policy before repeating the same steps.
	
	We use a three-layer ReLU network with 20 units in the $\pi_{G}$ input layer and the hidden layer. The input layer receives a state observation, while the output layer is a 9-unit action space, which assigns probabilities to each action using the softmax activation function. For $D_\psi$, we use a fully connected four-layer deep architecture. The input layer has 100 units, the first hidden layer has 50 units, and the second hidden layer has 20 units. The last layer outputs the probability using the sigmoid activation function. The value of the $D_\psi$ output is then used as the $R_\psi$ value to update the weights of both $D_\psi$ and $\pi_{G}$.

	\subsubsection{AV Standard Performance using $R_{\psi}$} \label{sec:threshold}
	
	Once we are able to learn $R_\psi$ using offline data and AIRL, we add an additional step of collecting $ R_{\psi} = (s,a) $ as part of our contribution to testing AVs. By doing so over several simulation episodes, we obtain a behavior representation of the AV under test to perform a few statistical analysis.

	\subsubsection{Statistical analysis of $R_{\psi}$ for finding threshold $ \beta $}
	
	By performing statistical analysis, we try to find confidence intervals to locate a minimum threshold $ \beta $ per scenario. This threshold value will determine which state-action pairs are found in uncertain action behaviors of AVs.

	\textit{Frequency-based distribution} is a statistical concept that refers to the occurrence of values in a given real-world dataset. While normal or known distributions follow a specific pattern or shape, frequency-based distributions do not have any distinct pattern or shape. When dealing with a frequency distribution, it is often useful to calculate the mean $ \mu $, standard deviation $ \sigma $, and confidence interval. 
	In our case, we use approximately 95\% of the data that fall within two standard deviations of the mean, whether the reward behavior distribution is normal or skewed. Instead of using both sides of the interval, we are interested in the left side of the distribution to calculate the $ \beta $ values per scenario. This can be denoted as:
	
	\begin{equation}
	\beta = \mu - 2\sigma  
	\end{equation}
	where $ \beta $ represents the threshold per reward distribution of a scenario. This threshold is utilized in the next and final step of the methodology to find the most likely failure events out of the normal-driven AV under test.

	\subsection{Step 3: Simulation testing using $R_\psi$}\label{sec:Step3}
	
	In the third and final step of our framework, we perform simulation tests using $R_\psi$ and $\beta$ to find the probability of failure states. As part of our framework, we propose to perform testing by first identifying the states that will most likely cause the AV to fail. Doing so strategically reduces the search space to uncertain events for existing AVs and then we can perform perturbation attacks as explained in Section~\ref{sec:disturbance}. The steps performed per scenario can be shown as the Algorithm~\ref{alg:2}.

	\begin{algorithm} \caption{ReMAV Simulation testing using $R_\psi$ and $\beta$}\label{alg:2}
		\begin{algorithmic}[1]
			\Require Set of trajectories $D = {s,a,r}$.
			\State Initialize an empty set of trajectories $D = \{\}$.
			\ForEach {episode} 
			\State Initialize an empty trajectory $\tau = [(s_1, a_1)]$.
			\State Observe the initial state $s_1$.
			\State Select an action $a_t$ according to the policy $\pi_{AV}(a_t|s_t)$.
			\State Observe the reward using $R_{\psi t}$ to obtain the reward $r_{\psi t}$.
			\If{$r_{\psi t}$ $<$ $\beta$}
			\State apply noise disturbance \{$\epsilon_1$, $\epsilon_2$\}.
			\EndIf
			\State Append the state-action pair $(s_t, a_t,r_{\psi t})$ to the trajectory $\tau$.
			\State Append the trajectory $\tau$ to the set of trajectories $D$.
			\EndFor
			\State Repeat for each timestep $t$ until the end of the episode.
			\State Return the set of trajectories $D$.

		\end{algorithmic}
	\end{algorithm}	
	
	\subsubsection{Disturbance Model}\label{sec:disturbance}
	
		ReMAV's main motivation is to demonstrate that once we have an overview of the standard AV driving behavior and its uncertain driving states, we can start with minimal noise attacks instead of learning complex adversarial attacking methods.
	\paragraph{Input Sensor Noise Perturbation}			
	One approach to generating noise perturbation to attack AVs in uncertain state events is using the Gaussian distribution. Our goal is to use this disturbance model to add minimal sampled noise to the AV's input image where the behavior of the AV is performed below a desired threshold.

	When adding Gaussian noise as an adversarial attack to an AV under test, it is imperative to take a strategic approach. Rather than randomly adding noise throughout the episode, a more effective approach is to selectively add noise based on the uncertainty of the self-driving car. This methodology can make a significant difference in the results. Randomly adding noise throughout the episode may not be as effective in deceiving the model, as it may not be targeted toward the specific areas of the input that the model is most sensitive to. However, selectively adding noise based on the uncertainty of the self-driving car can be more effective, as it can target the areas of the input that the model is most uncertain about. For example, if the model is uncertain about the distance of an object in front of the car, adding noise to that specific area can cause the model to make incorrect decisions, such as braking too early or too late. However, if noise is added randomly across the entire episode, the model may not be as affected as it may not be sensitive to all areas equally.

	To generate Gaussian noise, a random number generator is used to generate a set of values following a normal distribution. 
	The generated noise values can then be added to the sensor readings or control signals of the autonomous agent system, causing it to make incorrect decisions or take inappropriate actions. One advantage of using Gaussian noise to attack autonomous agents is that it can simulate realistic noise conditions that the system may encounter in its environment. Additionally, Gaussian noise is additive and can be easily incorporated into the system without requiring knowledge of the system's internal workings. 
	The probability density function of the Gaussian distribution for sampling noise $\epsilon_1$ is given by:
	
	$$ \epsilon_1 \sim \mathcal{N}(\mu, \sigma^2)$$
	
	where $\mu$ is the mean and $\sigma$ is the standard deviation. The attacker can use this distribution to generate noise by sampling its values. The amount of noise injected into the agent's perception or action space is controlled by the standard deviation of the distribution. In our experiments, we keep $\sigma$=0.001 in order to have a minimal perturbation under uncertain states.
	
	In order to test the robustness of an AV agent, it is often necessary to introduce noise into the input data to simulate real-world conditions. Let $x_t$ be the three-dimensional input data for the AV agent and $y_t$ be the output decision made by the agent based on $x_t$. Let $\epsilon_1$ be the 3-dimensional noise added to $x_{t+1}$ resulting in $\hat{x}_{t+1}$ based on the reward value lower than the threshold value. Once the behavior of the AV under test crosses the threshold of $ \beta $, we can add noise to the image as:

	$$  \hat{x}_{t+1} = x_{t+1} + \epsilon_1 $$

	$\hat{x}_{t+1}$ should be such that the added noise $\epsilon_1$ is representative of real world conditions and does not significantly alter the decision made by the AV agent. 
	
	\paragraph{NPC Action Perturbation}
	
	Another way to add minimal perturbation noise attacks within the testing simulations is to add noise within the action outputs of the NPCs (pedestrians and driving agents) in a multi-agent scenario. The uniform distribution is another commonly used probability distribution for generating noise. It is a continuous probability distribution where the probability of an event is equally likely over the range of possible values. The probability density function of the uniform distribution for the sampling noise $\epsilon_2$ is given by:
	
	$$ \epsilon_2 =
	\begin{cases}
	\frac{1}{b-a} & \text{if } a \leq x \leq b\\
	0 & \text{otherwise}
	\end{cases} $$
	
	where $x$ is the random variable and $a$ and $b$ are the lower and upper bounds of the distribution, respectively. The attacker can use this distribution to generate noise by uniformly sampling values from the range $[a, b]$. The amount of noise injected into the agent's perception or action space is controlled by the range of the distribution. In our experiments, we keep $a$ = 0.0 and $b$ = 0.001 to have a minimal perturbation in uncertain states.

	Let $y_t$ be the action output of the NPC agent. Let $ \epsilon_2$ be the noise added to $y_{t+1}$ as $\hat{y}_{t+1}$. Once the behavior of the AV under test crosses the threshold of $ \beta $, we can add noise to the NPC actions as:
	
	$$  \hat{y}_{t+1} = y_{t+1}+ \epsilon_2 $$
	
	$\hat{y}_{t+1}$ should be such that the added noise $ \epsilon_2$ is representative of real world conditions and does not significantly alter the decision made by the autonomous agent (AV under test in our case).

	\section{EXPERIMENTAL EVALUATION}\label{sec:EXPERIMENTALEVALUATION}
	

	The experiments aim to demonstrate the effectiveness of the proposed framework for testing driving policies in a multi-agent car environment. To this end, we first train AV policies in multiple driving scenarios to collect their state-action trajectories dataset. Next, we design a reward modeling technique by implementing the AIRL algorithm in order to learn a reward model \textbf{$R_\psi$} that represents the behavior of the AV under test. The same reward function is passed through statistical analysis to find threshold values \textbf{$\beta$} under which we assume AV to behave with uncertainty. Lastly, we run simulation testing and find rare failure scenarios using our threshold to add perturbation using our disturbance model. For evaluation, we perform ReMAV's comparison with two baselines. Furthermore, we compare ReMAV with prior research based on Bayesian optimization and adaptive stress-based adversarial testing techniques.

	\subsection{Research Questions}\label{sec:RQSS}
	
	During our experimental results and analysis, we ask the following research questions:
	
	\textbf{RQ1:} Can we observe uncertain states by only observing the reward distribution of an AV?
	
	\textbf{RQ2:} Can our framework effectively create challenging and unknown scenarios compared to baselines for the same AVs for testing?
	
	\textbf{RQ3:} Does the reward function help reduce the search space for finding edge cases compared to the baseline approaches?

	\textbf{RQ4:} How does ReMAV perform compared to the existing AV testing frameworks?
	
	By first observing uncertain states with reward modeling in RQ1, we thoroughly compare ReMAV in RQ2 and RQ3 with baselines defined in Section~\ref{sec:Baselines}. Comparison with baselines is made for all six evaluation metrics defined in Section~\ref{sec:metrics}. As for RQ4, we additionally compare with two prior methods using three additional metrics such as training-testing efficiency, total infractions detected, and simulation steps to find the first failure event.

	\subsection{Evaluation	Metrics}\label{sec:metrics}
	We evaluate the driving performance of the AVs under test using the following metrics:
	
	\begin{enumerate}
		\item CV: rate of collision with another vehicle
		\item CO: rate of collision with any other road objects
		\item CP: rate of collision with any other pedestrians
		\item OS: rate of offroad steering errors
		\item TTFC: the time it takes to have the first collision
		\item TTFO: the time it takes to have the first offroad occurrences
	\end{enumerate}

	We evaluate the ReMAV's ability to detect failure states in AVs by comparing its performance under noise disturbances to its standard performance. We calculate the percentage of error for each test episode using four metrics (CV, CO, CP, and OS), with 1500 simulation steps per episode and 50 episodes per scenario. Then we computed the average error rate for each metric on all episodes. Additionally, we measure the time in seconds that it takes for the ReMAV to detect the first collision (TTFC) and the first offroad occurrence (TTFO) in each test episode.
	
	
	\subsection{Experimental Setup}
	We use Town 3 scenario provided by the Python Carla API~\cite{town} in our partially observable urban-based driving environment. 
	\subsubsection{Driving Scenarios} \label{sec:ExpSteps}
	
\begin{figure}[!t]
		\centering
		
		\includegraphics[width = 0.5\textwidth,height=0.30\textheight]{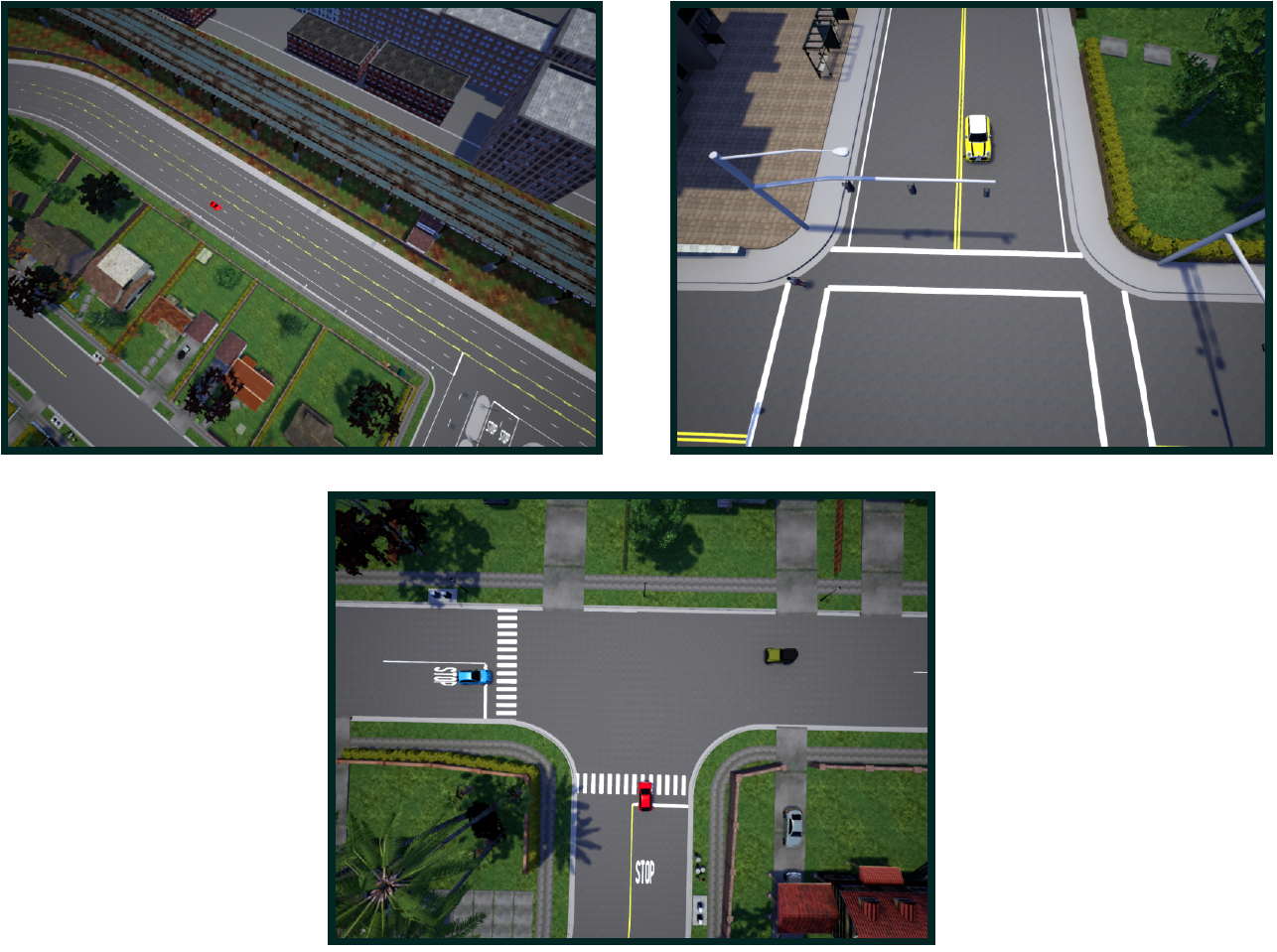}
		\caption{Illustration of Town03 Carla urban driving environment. The upper left subfigure represents the first driving scenario (Straight) and top right represents the second driving scenario (Pedestrian). Lastly, the bottom subfigure shows our third and final driving scenario (three-way) for the experimental evaluation of AVs under test.}
		\label{fig:Town03}
	\end{figure}

	Our experiments are configured for testing AVs in single- and multi-agent scenarios. The details of the training and testing configurations for our experiments are mentioned in Section~\ref{sec:ReMAVExperimentalSteps}. First, we provide brief descriptions of the driving scenario shown in Figure~\ref{fig:Town03}.
	
	\paragraph{Driving Scenarios 1 (Straight)}: We use a straight road from the Town03 map. The driving setting is simple but suitable, especially for validating single-agent AV policies in order to see their lane-keeping capability.
	
	\paragraph{Driving Scenarios 2 (Pedestrian)}: We use the pedestrian crossing scenario from the Town03 map. The driving setting is perfect for validating multi-agent AV policies in a scenario where pedestrians are faced.
	
	\paragraph{Driving Scenarios 3 (Three-Way)}: The scenario has independent non-communicating agents spawned close to the
three-way intersection throughout the testing scenarios. The choice of a three-way intersection as a driving scenario is based on its higher complexity for an AV agent.

	The three driving scenarios described here are good choices for evaluating the behavior of AVs due to their diversity and complexity. The straight scenario is simple but important, as driving in a straight line is a fundamental capability of any vehicle. The pedestrian scenario adds an additional level of complexity by introducing a dynamic agent that the autonomous vehicle must detect and respond to in real time. Finally, the three-way scenario is even more complex, as it involves multiple agents interacting with each other in a more dynamic environment. By evaluating autonomous vehicles in these diverse and challenging scenarios, we can gain a better understanding of their capabilities and limitations.
	
	\subsubsection{Comparisons with Baselines} \label{sec:Baselines}
	Since ReMAV is a carefully designed algorithm for the specific methodology, we established and implemented two comparison baselines for evaluation: a \textbf{R}andom \textbf{T}esting strategy (RT) and a \textbf{S}tress \textbf{T}esting strategy (ST). 
		These strategies allow us to thoroughly assess how AVs perform under random or continuous stress testing conditions, unlike our method, which only introduces noise perturbations when the AV is predicted to be in a state of uncertainty. Once we observe uncertain states using the reward distribution of an AV in RQ1, we compare ReMAV with RT and ST in RQ2 and RQ3 as mentioned in Section~\ref{sec:RQSS}.

	RT randomly selects a state within the testing episode to add image or NPC noise attacks. 
		This approach provides a more realistic assessment of an AV's performance in everyday scenarios, often characterized by random, unforeseen events. The steps performed by the RT baseline can be shown as an Algorithm~\ref{alg:3}.
	
	\begin{algorithm} \caption{Simulation testing using RT}\label{alg:3}
		\begin{algorithmic}[1]
			\Require Set of trajectories $D = {s,a,r}$.
			\State Initialize an empty set of trajectories $D = \{\}$.
			\ForEach {episode} 
			\State Initialize an empty trajectory $\tau = [(s_1, a_1)]$.
			\State Observe the initial state $s_1$.
			\State Select an action $a_t$ according to the policy $\pi_{AV}(a_t|s_t)$.
			\State Select a boolean value $rand = [(True, False)]$.
			\If{$rand == True$}
			\State apply noise disturbance \{$\epsilon_1$, $\epsilon_2$\}.
			\EndIf
			\State Append the state-action pair $(s_t, a_t)$ to the trajectory $\tau$.
			\State Append the trajectory $\tau$ to the set of trajectories $D$.
			\EndFor
			\State Repeat for each time-step $t$ until the end of the episode:
			\State Return the set of trajectories $D$.

		\end{algorithmic}
	\end{algorithm}

	ST, on the other hand, performs stress testing across all states of the episode using image or NPC noise attacks. Stress testing pushes an AV to its maximum using the same disturbance model to find any weak points. The steps performed by the ST baseline can be displayed as an Algorithm~\ref{alg:4}.
	
	\begin{algorithm} \caption{Simulation testing using ST}\label{alg:4}
		\begin{algorithmic}[1]
			\Require Set of trajectories $D = {s,a,r}$.
			\State Initialize an empty set of trajectories $D = \{\}$.
			\ForEach {episode} 
			\State Initialize an empty trajectory $\tau = [(s_1, a_1)]$.
			\State Observe the initial state $s_1$.
			\State Select an action $a_t$ according to the policy $\pi_{AV}(a_t|s_t)$.
			\State Apply noise disturbance \{$\epsilon_1$, $\epsilon_2$\}.
			\State Append the state-action pair $(s_t, a_t)$ to the trajectory $\tau$.
			\State Append the trajectory $\tau$ to the set of trajectories $D$.
			\EndFor
			\State Repeat for each time-step $t$ until the end of the episode:
			\State Return the set of trajectories $D$.

		\end{algorithmic}
	\end{algorithm}

	\subsubsection{Evaluation against Existing Frameworks} \label{sec:ComparisonDetails}
	
	We also compare our results with two well-known adversarial testing methodologies, namely BayesOpt~\cite{R2} and AST-BA~\cite{9636072}. 
	
	\textit{BayesOpt} is an adversarial testing approach based on a Bayesian optimization algorithm that attacks AVs by drawing physical lines on the road. BayesOpt simultaneously looks for failure events per training iteration to adversarially optimize its algorithm for the best solution. AST~\cite{R34} on the other hand, is an adaptive stress testing formulation that uses the MDP approach to find failures by attacking the simulation environment AV using RL in a low-fidelity environment. It then transfers its learning to a high-fidelity environment using the Backward algorithm (BA), which learns from expert demonstrations. AST-BA, similar to BayesOpt, trains and finds failure scenarios per training iteration. AST-BA has been evaluated on the Nvidia Drivesim urban driving simulator. On the contrary, ReMAV first trains on offline trajectories without the notion of adversarial attacks. Then, it uses the behavior representation to find failure events.

	\paragraph{Rationale behind the comparison and choices of algorithms}
	
	We selected the mentioned algorithms for comparison on the basis of the following rationale.
	
	\begin{enumerate}
		\item Both algorithms were tested in similar vision-based urban driving environments, such as Carla and Drivesim.
		\item Important evaluation metrics are included for comparison along with an in-depth analysis of other hyperparameters involved.
	\end{enumerate}
	
	While there have been numerous testing methodologies proposed in recent years for AV validation (as mentioned later in Section~\ref{sec:relatedwork}), we are constrained by certain technical limitations discussed in~\ref{sec:Conclusion}, which hinder an open comparison. Additionally, the lack of standard benchmarks also makes it challenging to select specific testing frameworks as state-of-the-art.

	\subsection{ReMAV Experimental Steps}\label{sec:ReMAVExperimentalSteps}

	\textbf{Step 1: Training AV Hyperparameters and Offline Data Collection}
	
	The details of the hyperparameters selected for the training of the AV driving agents are given in Table~\ref{tab:HyperparametersPPO}. The bottom rows of the table also provide information about offline trajectories collected once the AV standard behavior is available.
	
	\begin{table}[htbp]
		\caption{Hyperparameters for the PPO Based AV model.} \label{tab:HyperparametersPPO} 
		
		\begin{center}

			\resizebox{!}{3.0cm}{

				\begin{tabular}{lll}
					
					\toprule
					
					\textbf{Stage} & \textbf{Hyperparameter} & \textbf{Value} \\ \midrule

					\multirow{3}{*}{Replay Buffer} & Minibatch Range & 64 \\  
					
					& Epochs per Minibatch & 8 \\ 
					
					& Batch Mode & Complete Episodes \\ 
					
					\midrule
					\multirow{6}{*}{Updating AV Policy} & Discount factor ($\gamma$) & 0.99 \\ 
					
					& Clipping ($\epsilon$)  & 0.3 \\ 
					& KL Target & 0.03 \\ 
					& KL initialization & 0.3 \\ 
					& Entropy Regularizer & 0.01 \\ 
					& Value Loss Coefficient   & 1.0\\
					\midrule

					\multirow{5}{*}{Other AV Training Hyparameters }  & Total Training Steps & 1002016 \\ 
					& Total Training Episodes & 910  \\ 
					& Learning Rate & 0.0006  \\
					&	Batch Size & 128  \\ 
					& Optimizer	& Adam~\cite{R59}   \\

					\midrule \midrule
					\multirow{5}{*}{Offline Trajectories using AV}  & Total Driving scenarios	& 3   \\

					& Number of episodes & 150 (50 per scenario)  \\ 
					
					& Number of steps per episode & 2000 \\

					& State space dimensions & (12800,)  \\ 
					
					& Action space dimensions & (8,)  \\  
					
					& Trajectory collection cost & 4 hours  \\  
					\bottomrule

				\end{tabular}

			}

		\end{center}

	\end{table}
	
\begin{figure}[!t]
		\captionsetup[subfigure]{labelformat=empty}
		\subfloat[\footnotesize{(a)}]{%
			\includegraphics[, width=0.24\textwidth]{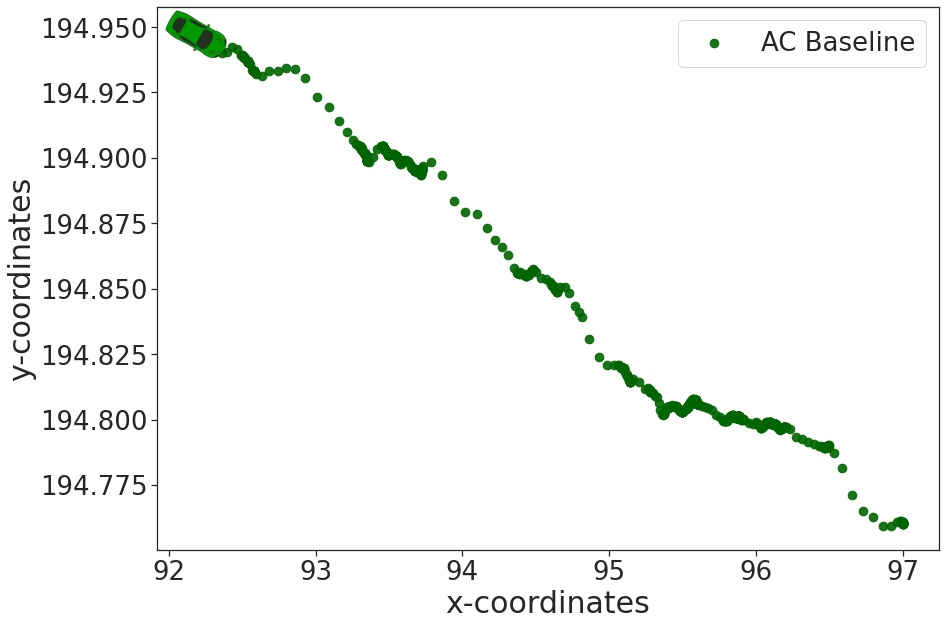}}
		\subfloat[\footnotesize{(b)}]{%
			\includegraphics[width=0.24\textwidth]{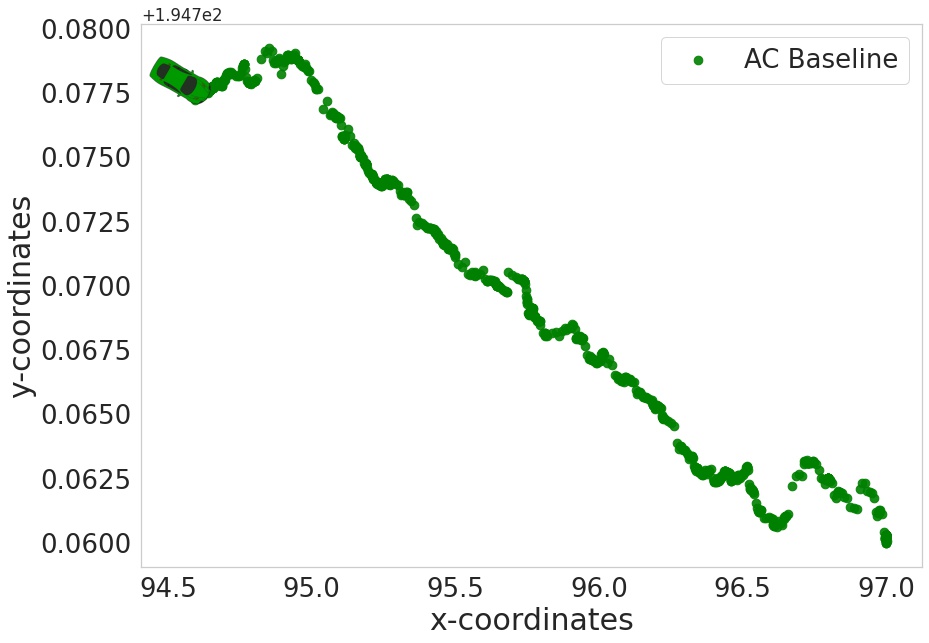}}
		\hspace{\fill}
		\centering
		\text{\small{Straight}}
		\subfloat[\footnotesize{(c)}]{%
			\includegraphics[ width=0.24\textwidth]{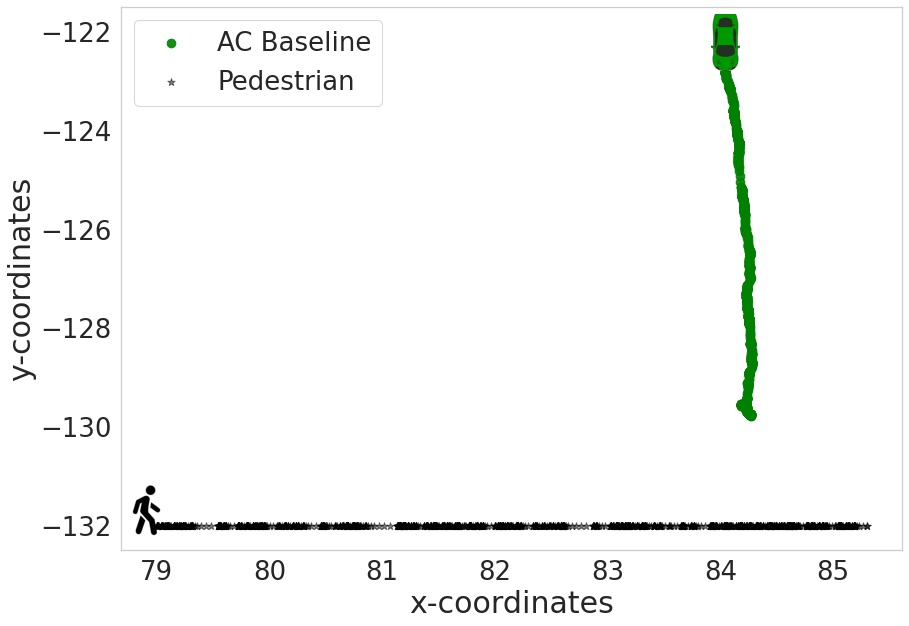}}
		\subfloat[\footnotesize{(d)}]{%
			\includegraphics[ width=0.24\textwidth]{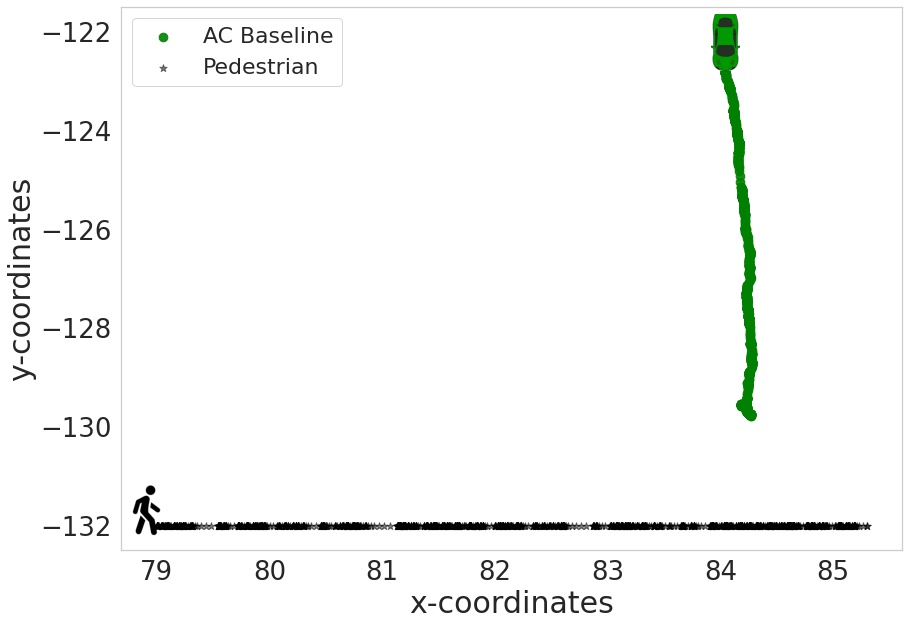}}
		\hspace{\fill}
		\centering
		\text{\small{Pedestrian}}
		\subfloat[\footnotesize{(f)}]{%
			\includegraphics[ width=0.24\textwidth]{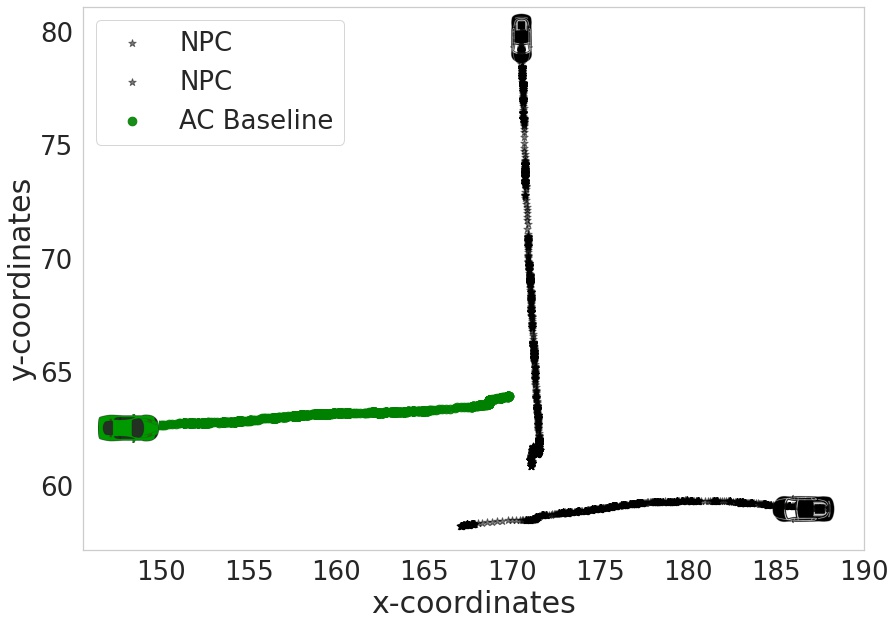}}
		\subfloat[\footnotesize{(g)}]{%
			\includegraphics[ width=0.24\textwidth]{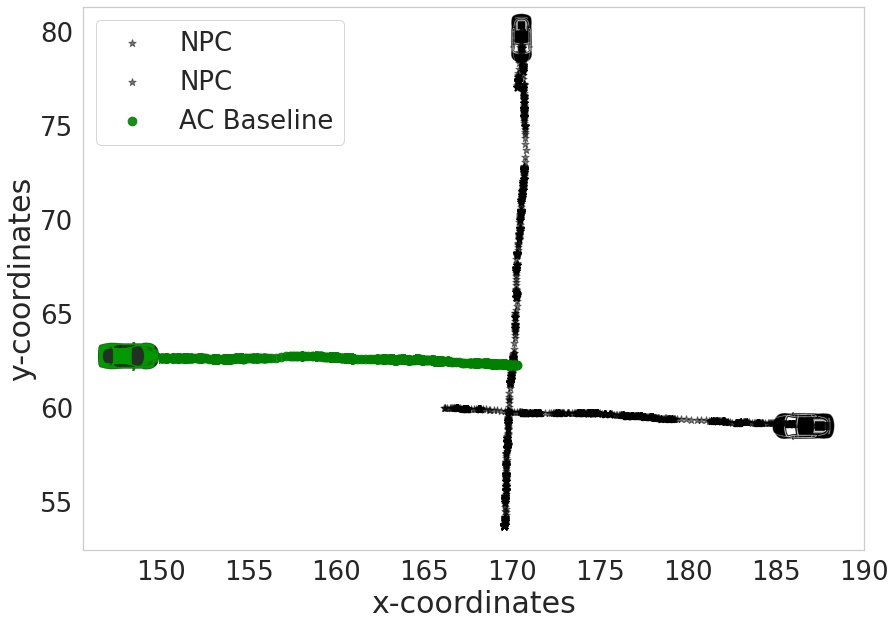}}
		\hspace{\fill}
		\centering
		\text{\small{Three-way}}
		\caption{\label{fig:Baseline} 2D visualization of the AV standard driving coordinates in all three scenarios.}
	\end{figure}

	Standard AV behavior can be visualized in a top-down view using Figure~\ref{fig:Baseline}. Each row shows episodic examples of the AV in three driving scenarios. Their behavior with respect to the evaluation metrics is discussed in depth in Section~\ref{sec:results}.

	\textbf{Step 2: Training AIRL Hyperparameters and Gathering AV Behavior Data}

	Once we have offline data of state-action pairs from the first step of ReMAV, we perform the AIRL algorithm to train $R_\psi$ as part of the discriminator network. The details of the AIRL algorithm can be seen in Table~\ref{tab:AIRLHyperparameters}. The bottom rows of the table represent the part where we utilize $R_\psi$ to obtain AV behavior and their reward values.
	
	\begin{table}[htbp]
		\caption{Hyperparameters for the AIRL algorithm.} \label{tab:AIRLHyperparameters} 
		
		\begin{center}

			\resizebox{!}{2.8cm}{

				\begin{tabular}{lll}
					
					\toprule
					
					\textbf{Stage} & \textbf{Hyperparameter} & \textbf{Value} \\ \midrule
					
					\multirow{5}{*}{Updating G Policy}   
					
					& gen\_replay\_buffer\_capacity & 128 \\


					& Clipping ($\epsilon$)  & 0.2 \\ 
					& KL Target & 0.03 \\ 
					& KL initialization & 0.3 \\ 
					& Entropy Regularizer & 0.01 \\ 
					\midrule 
					
					\multirow{3}{*}{Updating D Policy} & Discount factor ($\gamma$) & 0.99 \\ 
					
					& n\_disc\_updates\_per\_round   &  4 \\
					& Training Frequency & 32 \\

					\midrule 
					
					\multirow{5}{*}{Other Hyparameters }  &	Number of steps per episode & 2000 \\ 
					
					& Total Training Episodes & 500  \\ 
					& Learning Rate & 0.0006  \\
					&	Batch Size & 128  \\ 
					& Optimizer	& Adam~\cite{R59}   \\
					
					\midrule \midrule
					\multirow{6}{*}{Collecting AV Performance Using $R_\psi$}  & Total Driving scenarios	& 3   \\

					& Number of episodes & 60 (20 per scenario)  \\ 
					
					& Number of steps per episode & 2000 \\

					& State space dimensions & (12800,)  \\ 
					
					& Action space dimensions & (8,)  \\  
					
					& Reward dimensions & (1,)  \\  
					
					\bottomrule
					
				\end{tabular}

			}

		\end{center}

	\end{table}

\begin{figure}[!b]
		\centering
		
		\includegraphics[width = 0.52\textwidth,height=0.29\textheight]{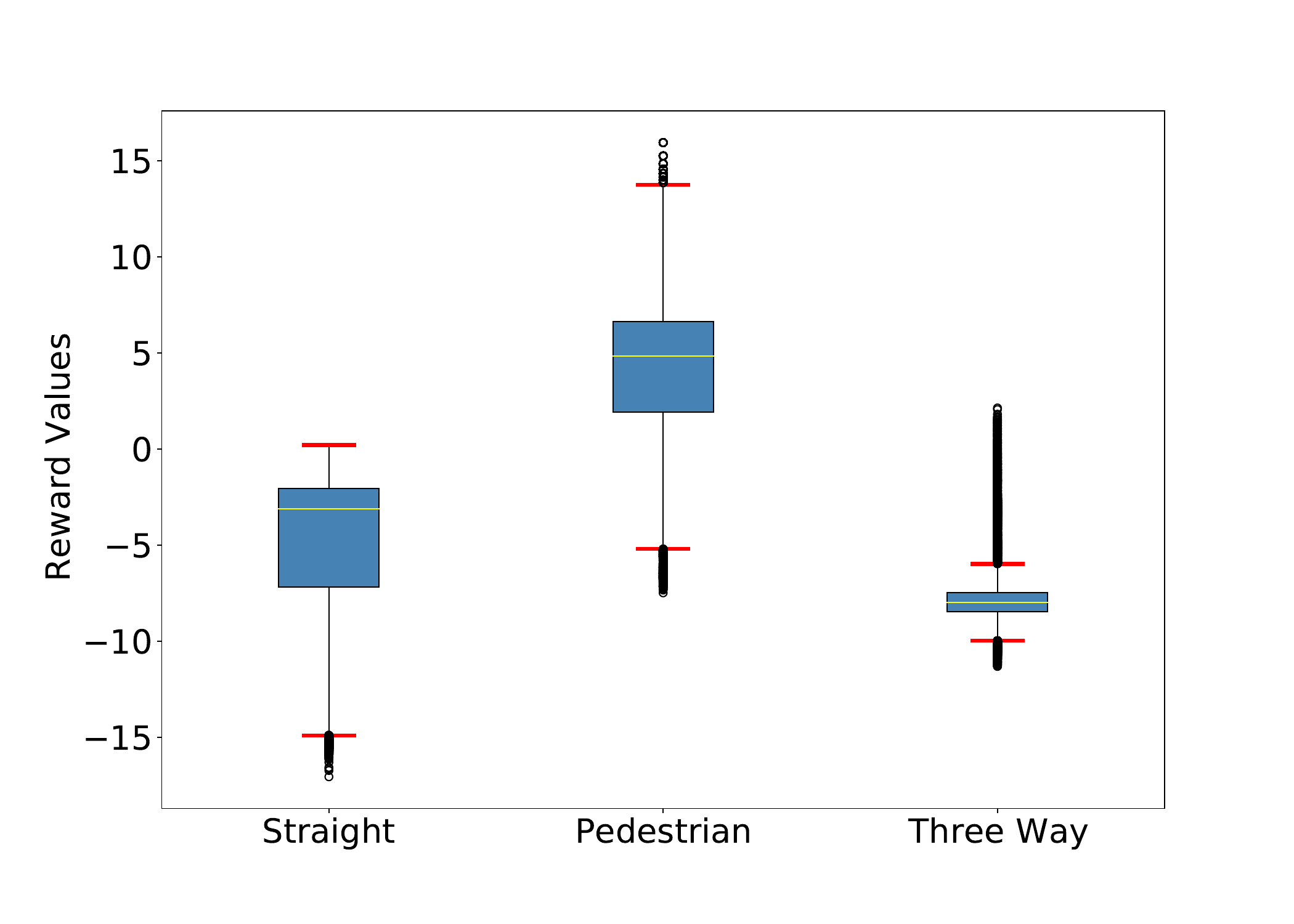}
		\caption{Boxplot for $R_\psi$ value distribution across three driving scenarios. The red lines in the boxplots represent the minimum and maximum expected values beyond which we see some outliers' reward values. The standard AV driving model performs comparatively better in a pedestrian scenario with a positive mean above $R_{\psi}$ values. }
		\label{fig:boxplot}
	\end{figure}

	Figure~\ref{fig:boxplot} represents the distribution of $R_\psi$ per each driving scenario for standard AV performance. Overall, AV performs better in the pedestrian scenario, while its actions are the least confident in the three-way scenario.
	
	Once we have the data with the values $R_\psi$ from the previous step as standard AV, we perform the third step of the framework to first obtain $\beta$ for each scenario and use it for simulation testing under selective disturbance perturbations.

	\textbf{Step 3: Hyperparameters for Simulation Testing using $R_\psi$ and $ \beta $ }

 Table~\ref{tab:TestingHyperparameters} provides the details of the number of episodes used in testing image or NPC action-based perturbations, as well as the number of steps per episode.

	\begin{table}[htbp]
		\caption{Hyperparameters for the Testing Phase in ReMAV} \label{tab:TestingHyperparameters} 
		
		\begin{center}
			\resizebox{!}{0.8cm}{
				\begin{tabular}{ll}
					
					\toprule
					
					\textbf{Hyperparameter} & \textbf{Value}  \\ \midrule
					
					Total Testing Episodes for Image Noise Testing per scenario & 50  \\ 
					Total Testing Episodes for NPC Action Perturbation  per scenario & 50  \\ 
					Number of steps per episode & 1500 \\ 
					
					\bottomrule
					
				\end{tabular}
			}
		\end{center}
	\end{table}
	
	\subsection{Simulation Setup}
	
	We utilize~\textit{RLlib}~\cite{R54} from the Ray framework to implement advanced AV driving policies. Our training, testing, and validation of AVs are performed using the urban driving simulation framework \textit{Carla}~\cite{R51}. Both RLlib and Carla are integrated by an existing open source platform~\cite{10062456} which also provides OpenAI's Gym toolkit for a multi-agent urban driving environment. Our DRL-based model architectures are created using \textit{TensorFlow}~\cite{R53} version 2.1.0, as part of the RLlib library.

	\section{Results and Analysis}\label{sec:results}
	In this section, we analyze the results collected in Section~\ref{sec:EXPERIMENTALEVALUATION} using our desired research questions.

	\subsection{RQ1: Can we observe uncertain states by only observing the reward distribution of an AV?}
	
	To observe uncertain regions, we look at the reward values predicted using $R_\psi$ for every pair of state-actions in simulation. This will be considered standard behavior for AV before we compare it with testing under noise perturbations. 
	
\begin{figure*}[!ht]
		\captionsetup[subfigure]{labelformat=empty}
		\subfloat[\footnotesize{(a)}]{%
			\includegraphics[, width=0.34\textwidth]{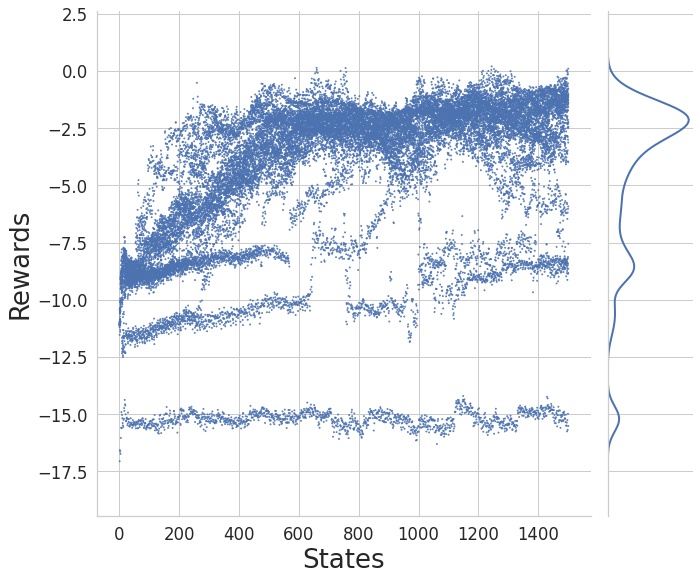}}
		\subfloat[\footnotesize{(b)}]{%
			\includegraphics[width=0.34\textwidth]{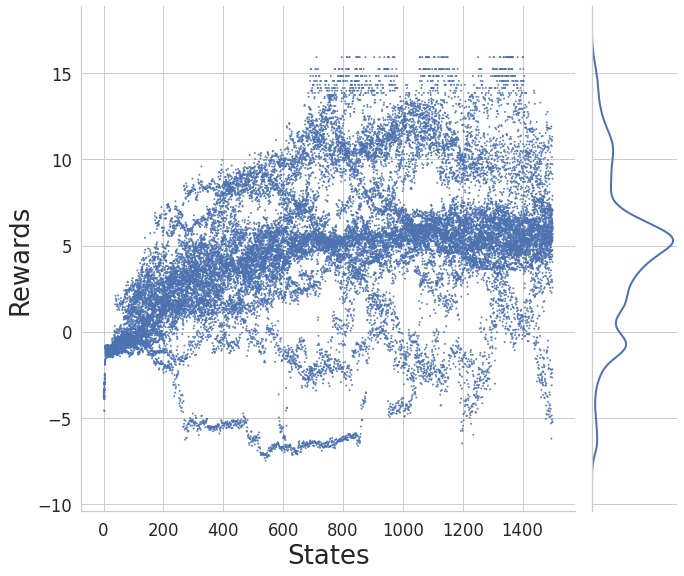}}
		\subfloat[\footnotesize{(c)}]{%
			\includegraphics[ width=0.34\textwidth]{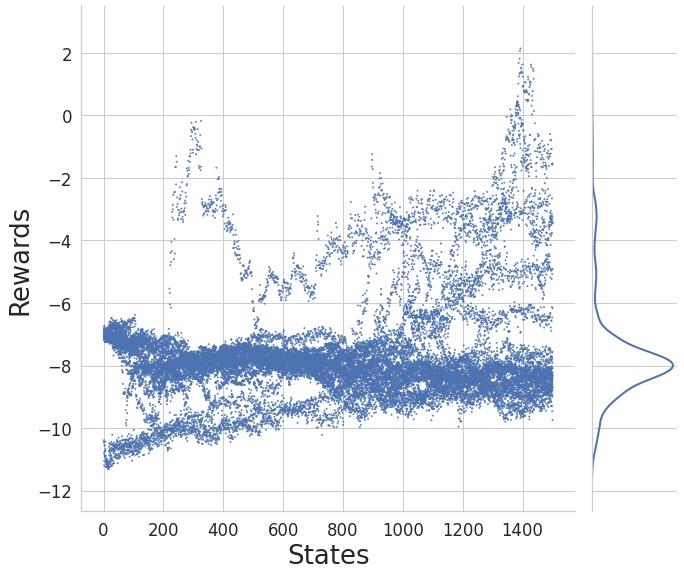}}
		
		\hspace{\fill}
		
		\subfloat[\footnotesize{(d)}]{%
			\includegraphics[, width=0.34\textwidth]{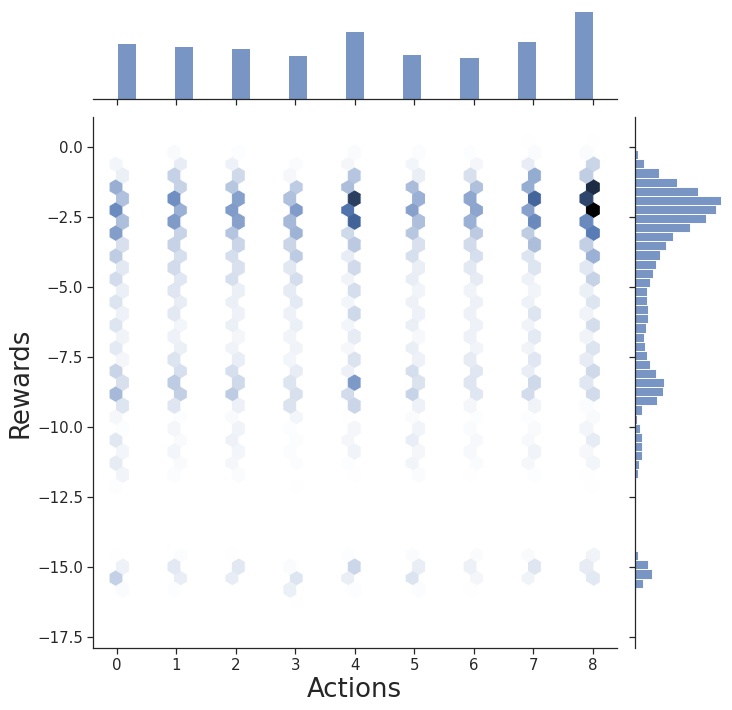}}
		\subfloat[\footnotesize{(e)}]{%
			\includegraphics[width=0.34\textwidth]{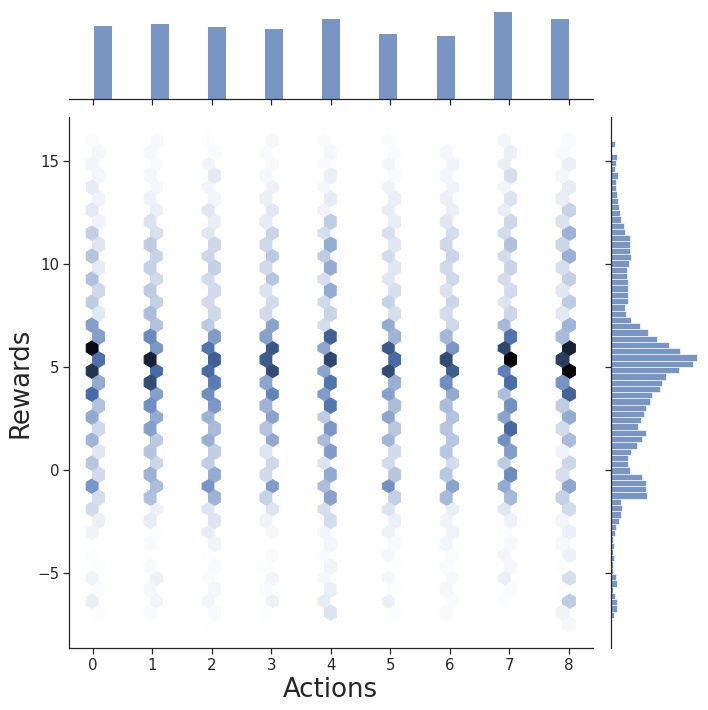}}
		\subfloat[\footnotesize{(f)}]{%
			\includegraphics[ width=0.34\textwidth]{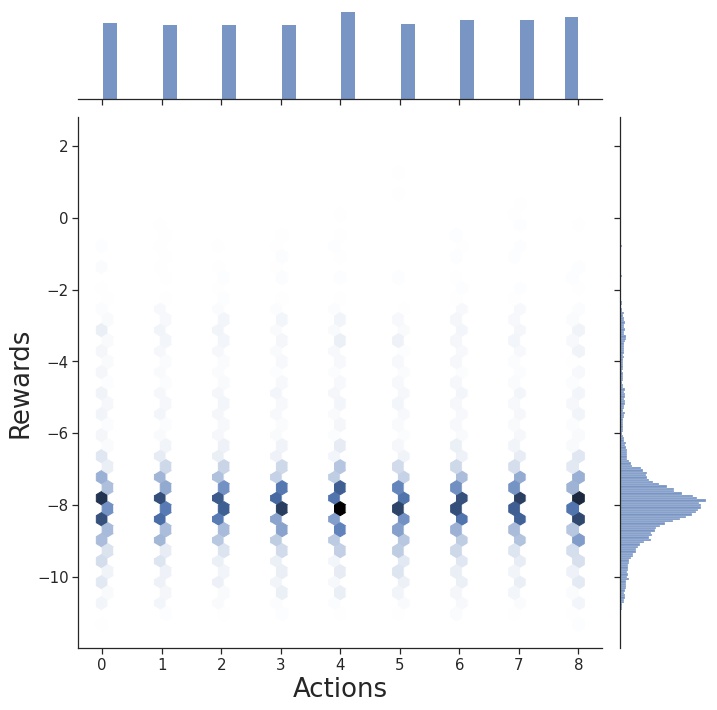}}
		\hspace{\fill}
		\subfloat{\includegraphics[width=0.34\textwidth]{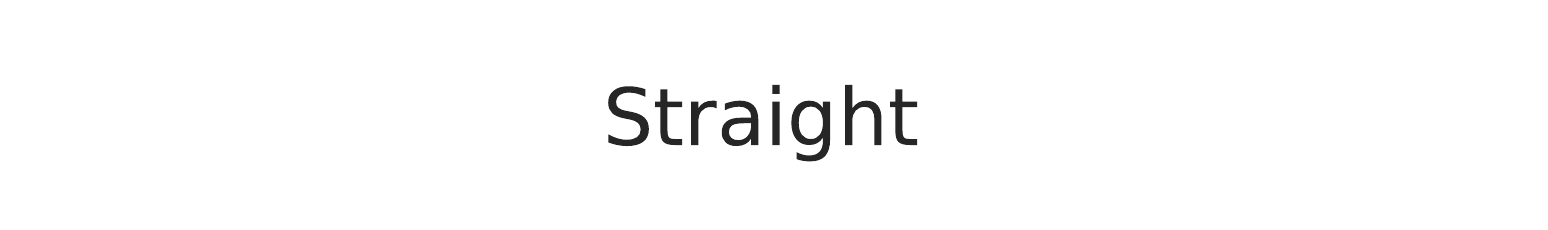}}
		\subfloat{\includegraphics[width=0.34\textwidth]{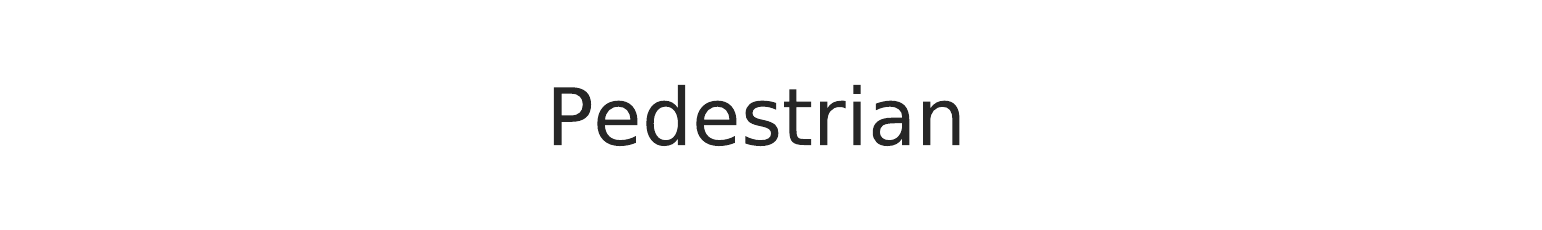}}
		\subfloat{\includegraphics[width=0.34\textwidth]{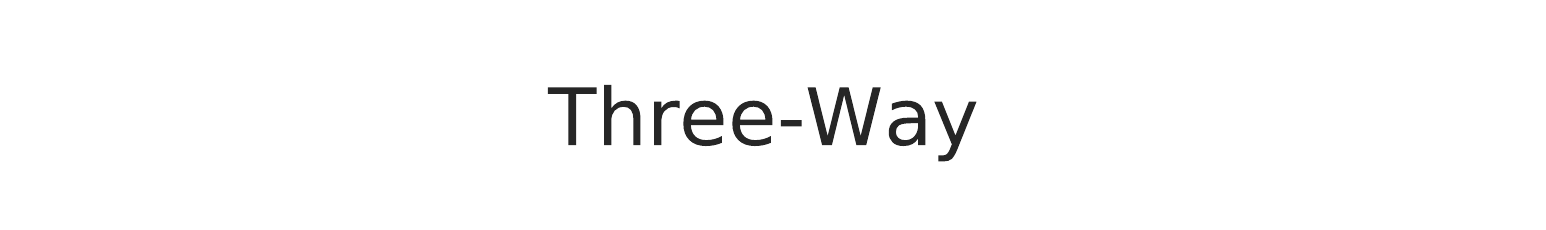}}
		
		\caption{\label{fig:RQ1_1} Multivariate graph representation of the reward distributions of $R_{\psi}$ for the state distributions (top row) and action distributions (bottom row). (a) (d) is directed towards AV driving performance in a straight scenario, (b) (e) for pedestrians, and (c) (f) for a three-way. (a) and (b) show standard positive AV behavior that shows confident driving decisions compared to (c) where the reward values lie mostly in a negative direction while visiting states. We observe some outlier reward values in a negative direction in (a) and (b) scenarios, while it is hard to observe such outliers when visualizing (c).	In the bottom row, we see an equal distribution of actions in (e) and (f) contributing to AV overall behavior as opposed to in (d) where only a few actions are taken the most throughout the single-agent straight scenario. }
	\end{figure*}

	Figure~\ref{fig:RQ1_1} represents multivariate graphs for states with reward and actions with reward, respectively. The graphs for all three scenarios between state and reward describe the overall driving behavior of the AV and the regions of interest where they were either confident or uncertain. From the top row graphs in the figure, we can clearly see some states for straight driving scenarios where $R_\psi$ gave the lowest reward values. For the pedestrian scenario, we see good driving conditions with only a few states visually lying as outliers. As for the state-reward representation for a three-way scenario, the driving behavior distribution mainly lies more on the negative-reward end, giving us a hint of the most likely failures we can get.
	
	The bottom row of Figure~\ref{fig:RQ1_1} shows the reward values of the standard AV model concerning the action space. We can see a similar behavior as in the state-reward representation shown in the top row of the figure for all scenarios. For a straight scenario, only a few actions influenced more in the driving decision-making of AV while in the pedestrian and three-way scenario, we see that almost all of the actions in the discrete action space contribute to the AV driving behavior.

\begin{figure*}[!ht]
		\captionsetup[subfigure]{labelformat=empty}
		\subfloat{%
			\includegraphics[, width=0.34\textwidth]{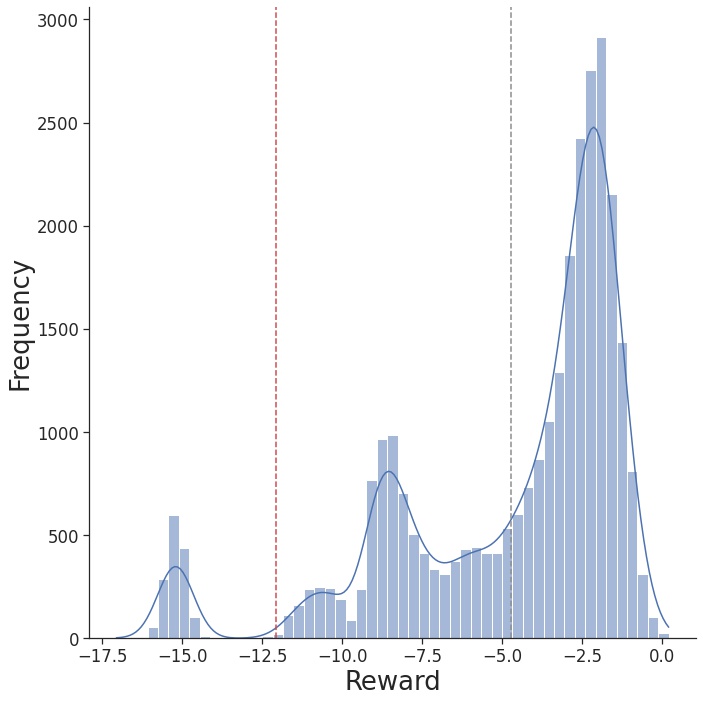}}
		\subfloat{%
			\includegraphics[width=0.34\textwidth]{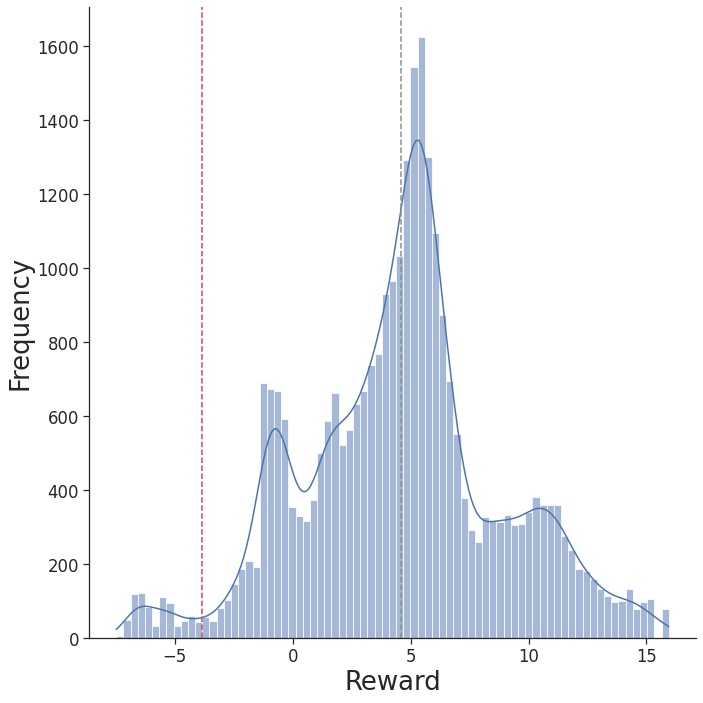}}
		\subfloat{%
			\includegraphics[ width=0.34\textwidth]{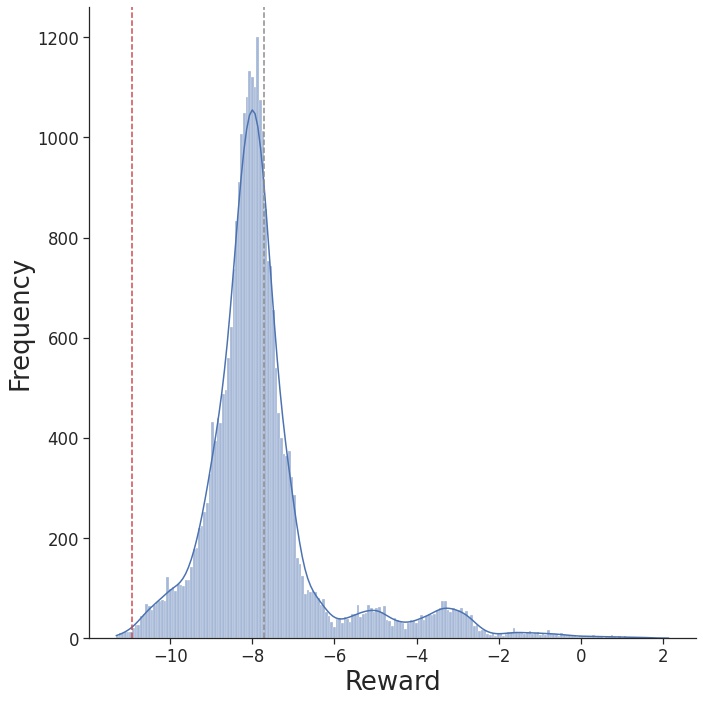}}
		\hspace{\fill}
		\subfloat{\includegraphics[width=0.348\textwidth]{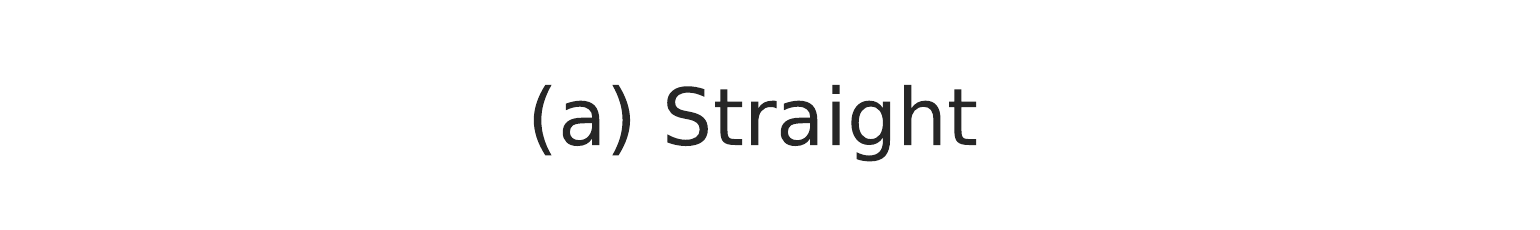}}
		\subfloat{\includegraphics[width=0.348\textwidth]{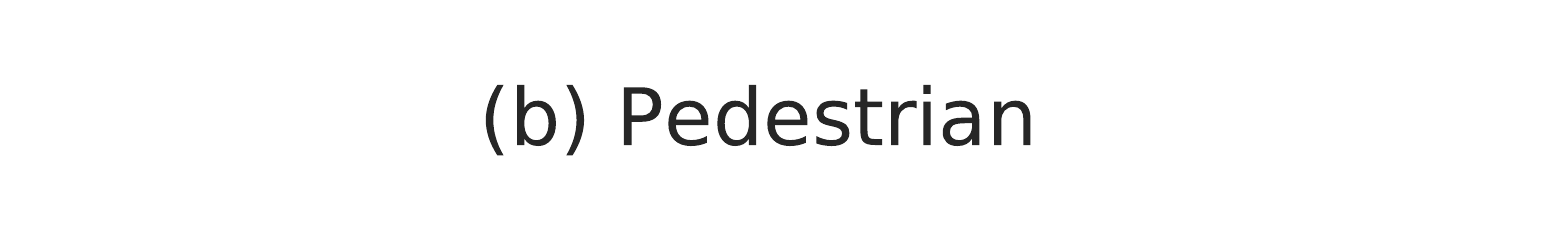}}
		\subfloat{\includegraphics[width=0.348\textwidth]{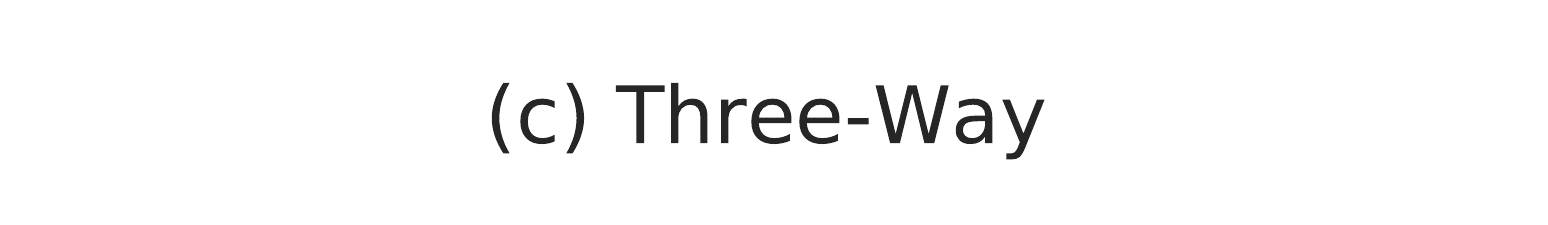}}
		\caption{\label{fig:RQ1_2} $R_{\psi}$ distribution in all driving scenarios to model standard AV performance. The red line represents the threshold $\beta$ for the simulation testing phase calculated using 95\% of the confidence interval. In straight, pedestrian, and three-way scenarios, we obtain $\beta$=[-12.05, -3.847, -10.92] respectively during our experiments. According to $R_{\psi}$, the standard AV has an overall decent performance in the straight (a) and pedestrian (b) scenarios compared to the three-way (c) scenario.}
	\end{figure*}

	As discussed in Section~\ref{sec:ReMAV}, our goal is to use the distribution $R_\psi$ and analyze it to define the threshold. A threshold $\beta$ will be different for all scenarios given the $R_\psi$ value calculated by given state-actions as input. The process can be illustrated in Figure~\ref{fig:RQ1_2} where the graphs represent the distribution $R_\psi$ against the states-action pairs as AV. $R_\psi$ in a straight environment is skewed in the right direction, and the driving behavior of AV as a single agent is very promising. For the pedestrian scenario shown in the middle figure, we can clearly see that the distribution is close to the normal distribution. As for the three-way, it is more skewed toward the left values of the $R_\psi$, showing that the threshold might just not cover many areas, but we might be able to find failure scenarios.
	
	The vertical red line in each of the $R_\psi$ distributions in the figure describes our estimated threshold $\beta$ for all three scenarios. As described in Section~\ref{sec:threshold}, we use a confidence interval of 95\% and use the left boundary of the interval as $\beta$. In straight, pedestrian, and three-way scenarios, we obtain $\beta$=[-12.05, -3.847, -10.92] respectively within our experimental setup.
	
In summary, RQ1 discusses the observation of uncertain states in an AV by analyzing the reward distributions. Multivariate graphs are used to represent driving behavior and regions of interest. We also explain the process of defining a threshold $\beta$ based on the reward distribution and present the estimated thresholds for different scenarios.

	\subsection{RQ2: Can our framework effectively create challenging and unknown scenarios compared to baselines for the same AVs for testing?}

	After obtaining $\beta$ values in RQ1, our objective is to utilize them to answer RQ2. For our experiments, as mentioned in Step 3 of Section~\ref{sec:ReMAVExperimentalSteps}, we used 50 episodes to test the AVs.
	
	Table~\ref{tab:RQ2_1} provides a comparison of the standard behavior of the AV versus its performance under image noise and NPC action perturbations. We use 4 of the 6 evaluation metrics defined in~\ref{sec:metrics} where a value closer to 0 means a low percentage of error found, while values near 1 show more occurrences of failure events within the testing phase. For the straight scenario, we only use CO and OS as evaluation metrics since there are no pedestrian and driving NPCs involved, whereas, in a pedestrian scenario, we use CO, CP, and OS as metrics as we exclude CV due to the non-presence of NPC driving agents. In the end, three-way scenarios are evaluated using CV, CO, and OS with no pedestrian information to record as CP. 
	
\begin{figure*}[!ht]
		\captionsetup[subfigure]{labelformat=empty}
		\subfloat[\footnotesize{(a)}]{%
			\includegraphics[, width=0.34\textwidth]{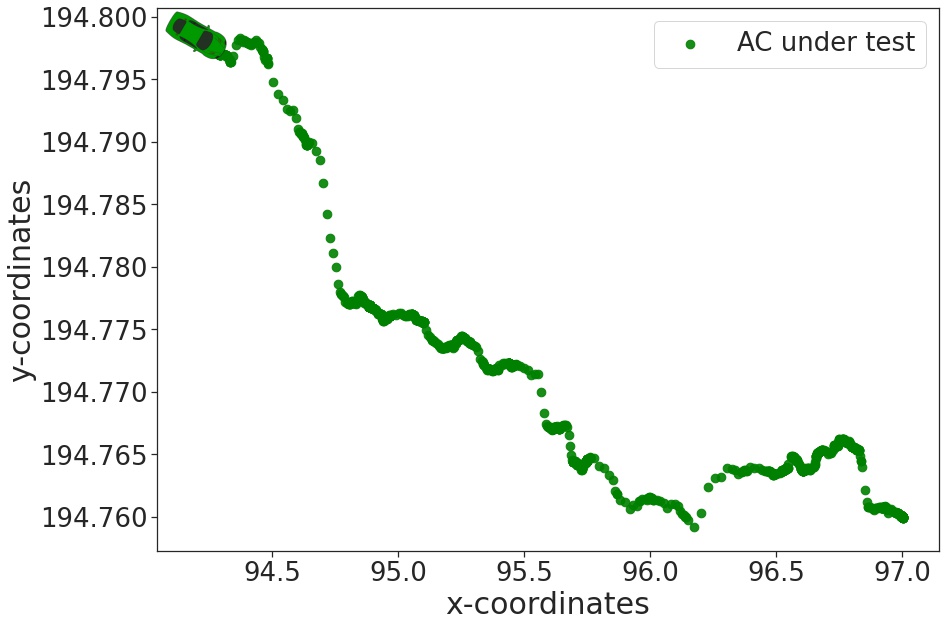}}
		\subfloat[\footnotesize{(b)}]{%
			\includegraphics[width=0.34\textwidth]{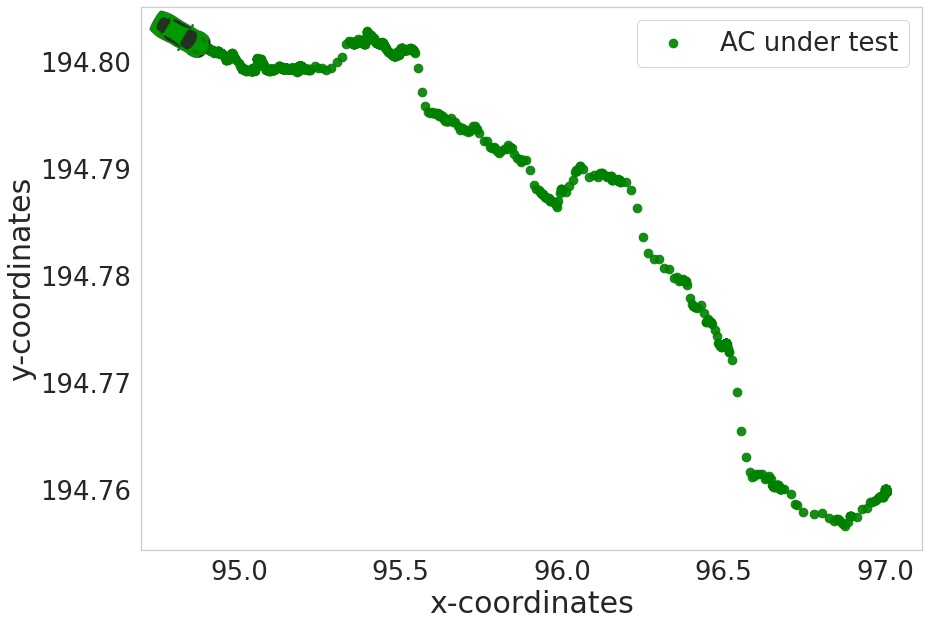}}
		\subfloat[\footnotesize{(c)}]{%
			\includegraphics[width=0.34\textwidth]{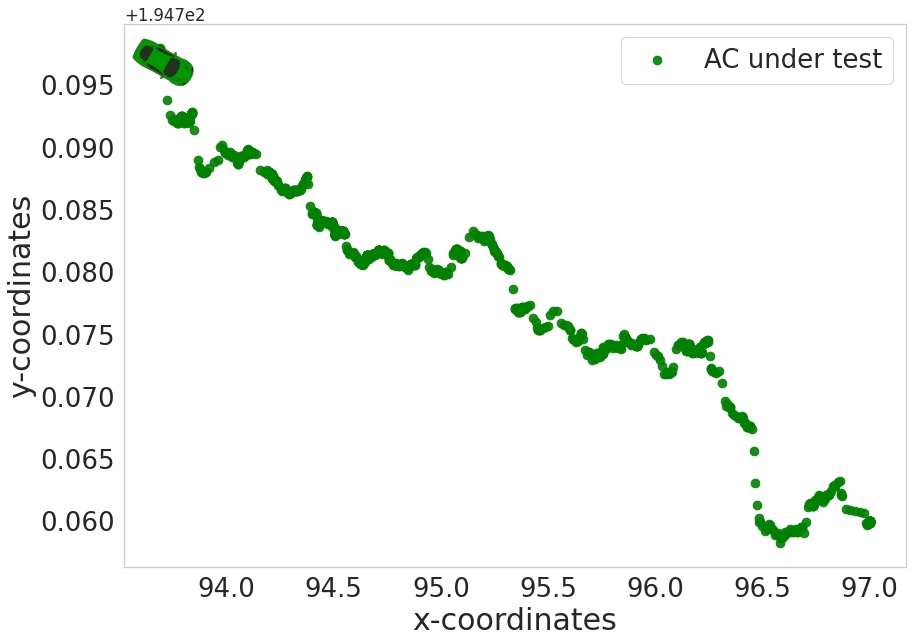}}
		\hspace{\fill}
		\subfloat[\footnotesize{(d)}]{%
			\includegraphics[width=0.34\textwidth]{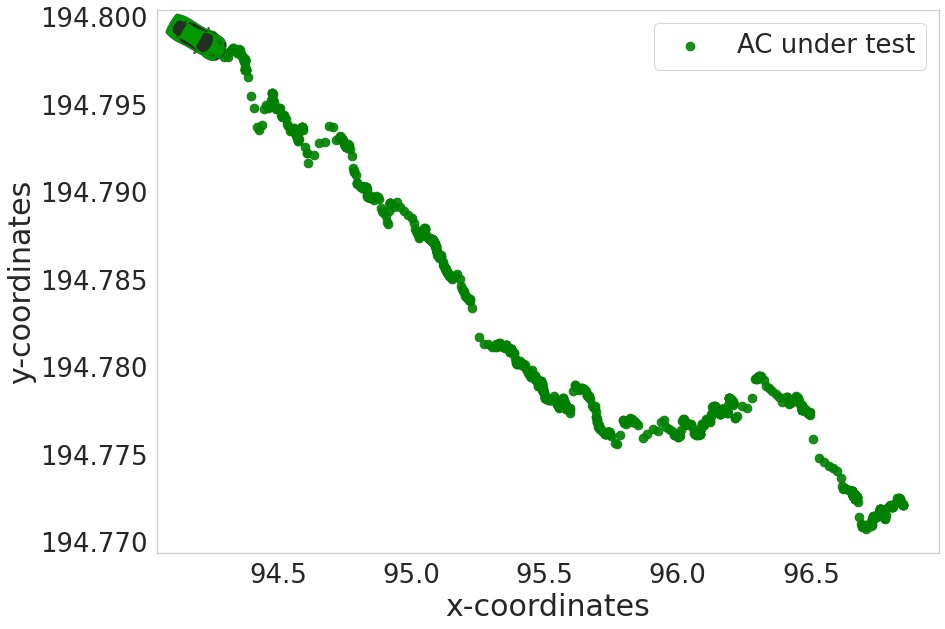}}
		\subfloat[\footnotesize{(e)}]{%
			\includegraphics[width=0.34\textwidth]{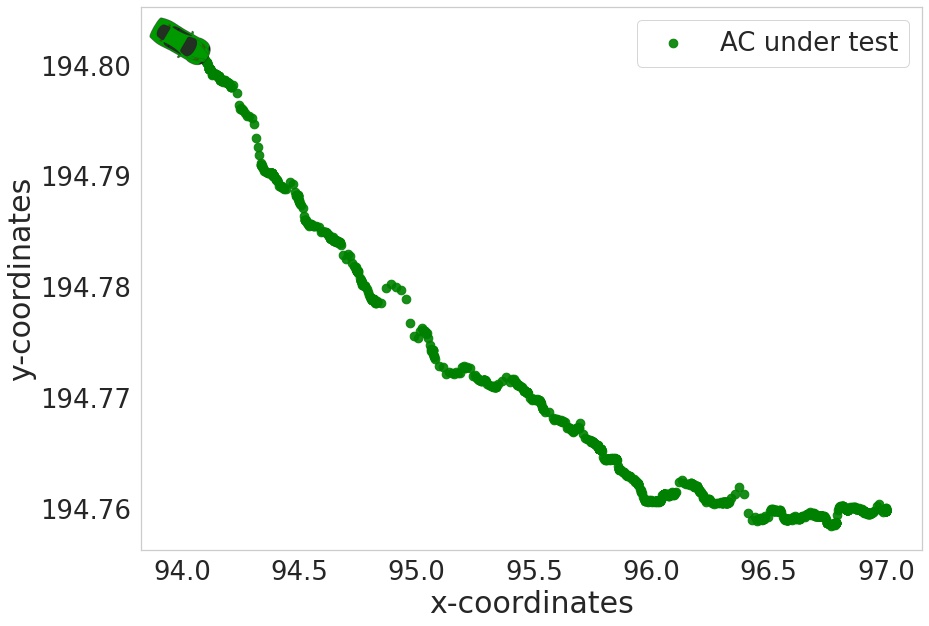}}
		\subfloat[\footnotesize{(f)}]{%
			\includegraphics[width=0.34\textwidth]{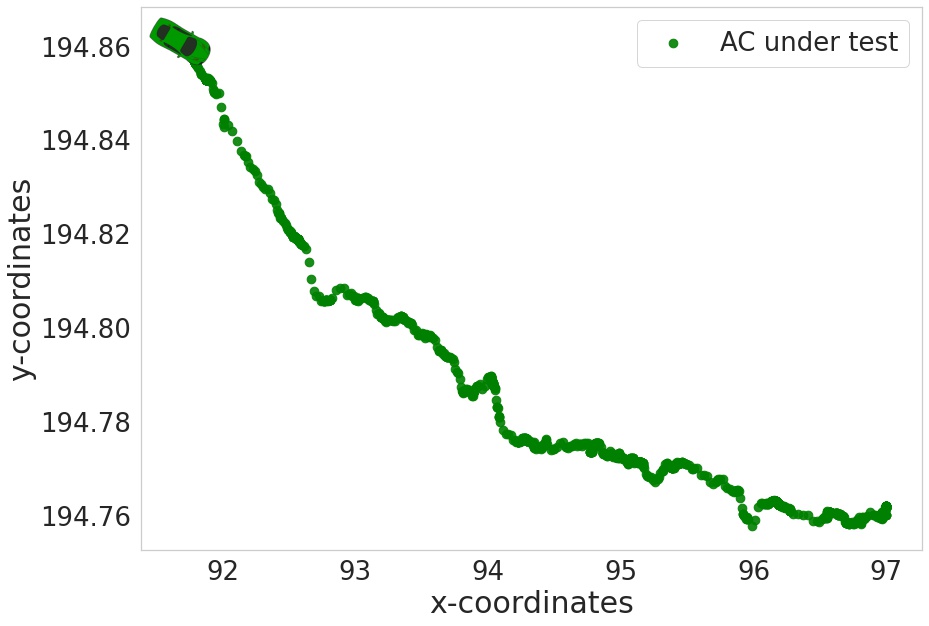}}

		\caption{\label{fig:RQ2_2} Top-down visualization of the AV driving coordinates while testing the AV agent in the presence of disturbance attacks. The testing is performed in a straight driving scenario. (c), (e), and (f) represent the perfect driving episodes where the AV under test was still able to perform well. Whereas we see offroad steering errors observed in either starting phase like (a) and (b) when faced with image perturbation attacks, but the AV agent was able to recover in the rest of the episodic run. We also observe some offroad steering instances as in (a) and (d) at the end of the testing episodes.
		}
	\end{figure*}

	As a standard performance without testing involved, the AV performs perfectly with no collision or offroad steering errors in the straight scenario. It almost performs similarly in pedestrian and three-way scenarios with rare collision and offroad steering occurrences. As we move towards the testing phase, we start by using minimal image noise perturbations with our disturbance model defined in Section~\ref{sec:disturbance}. RT and ST baseline failed to find any failure events in a straight scenario in the testing phase. As for the ReMAV, we can see that our disturbance attacks are not enough to force AV in a straight single-agent scenario even where we were able to identify few state-action representations below the required threshold. AV behaved well under test with no road collisions but faced a minute increase in the offraod steering errors. This shows that although the default AV driving appeared to have visible driving states with lower reward values, the overall performance of the AV using $R_\psi$ was positively distributed, resulting in better performance. We can visualize the performance of the AV under test in a straight scenario using Figure~\ref{fig:RQ2_2}. We can see that even under the influence of the image perturbations, the AV was able to recover at the start and end of the episodes, resulting in driving mostly straight with no collision on both sides of the road. It faces minor offroad steering errors using ReMAV but in a single agent situation, it was able to recover back to normal driving behavior hence performing mostly above the threshold value.
	
	\begin{table*}[!htbp]
		
		\begin{center}
			\caption{Comparison of the behavior of AV under test before and after adding noise in the testing phase, in terms of CV, CO, CP, and OS metrics, averaged across 50 episodes. Using ReMAV, AVs have more collisions and offroad steering errors in the presence of minimal disturbance attacks when their driving behavior is found uncertain.} \label{tab:RQ2_1}
			\resizebox{!}{1.5cm}{
				\begin{tabular}{ll|ll|lll|llll}
					
				& \multicolumn{1}{c|}{}	& \multicolumn{2}{c|}{Straight}   & \multicolumn{3}{c|}{Pedestrian} & \multicolumn{3}{c}{Three-Way} \\

				&	\multirow{4}{*}{}   & CO (\%)   & OS (\%)  &  CO (\%) &  CP (\%)  & OS (\%) & CV (\%) & CO (\%)   & OS (\%)    \\ 
					\midrule
					
				&	AV standard behavior & 0.0    & 0.0   & 0.0 & 0.008 $ \pm $ 0.0018 &  0.214 $ \pm $ 0.002  &  0.0542 $ \pm $ 0.0013& 0.0354  $ \pm $ 0.0010&  0.1887  $ \pm $  0.0022   \\ 
					\midrule

				\multirow{2}{*}{RT}
				&	AV under image noise attack $ \epsilon_1 $   & 0.0 & 0.0 & 0.0 & 0.0594 $\pm$ 0.0013 & 0.1 $\pm$ 0.002 & 0.129 $\pm$ 0.0019 & 0.0 & 0.192 $\pm$ 0.0022 \\ 
					
				&	AV under NPC noise attack $ \epsilon_2 $   & - & - & 0.04 $\pm$ 0.0012 & 0.13 $\pm$ 0.0028 & 0.19 $\pm$ 0.002 & 0.1112 $\pm$ 0.0018 & 0.0153 $\pm$ 0.0007 & 0.133 $\pm$ 0.0019  \\ 
					\midrule
				\multirow{2}{*}{ST}	&	AV under image noise attack $ \epsilon_1 $   & 0.0 & 0.0 & 0.0 & 0.068 $\pm$ 0.0014 & 0.1377 $\pm$ 0.0019 & 0.145 $\pm$ 0.002 & 0.0 & 0.123 $\pm$ 0.0019  \\ 
					
				&	AV under NPC noise attack $ \epsilon_2 $   & - & - & 0.0 & 0.2 $\pm$ 0.0026 & 0.0 & 0.0 & 0.0 & 0.01 $\pm$ 0.001  \\ 
					
					\midrule	
					
								\multirow{2}{*}{ReMAV}
					&	AV under image noise attack  $ \epsilon_1 $   &  0.0   & \textbf{0.12 $ \pm $ 0.018}   &\textbf{0.160 $ \pm $ 0.0004} &  \textbf{0.328 $ \pm $ 0.001}  & \textbf{0.341 $ \pm $ 0.0024}& \textbf{0.375 $ \pm $ 0.0021} & \textbf{0.249 $ \pm $ 0.0008}  & \textbf{0.69826 $ \pm $ 0.0022}  \\ 
					
					&	AV under NPC action attack $ \epsilon_2 $   &  -   & -   & \textbf{0.238 $ \pm $ 0.0012} &  \textbf{0.4883 $ \pm $ 0.0026}   & \textbf{0.3428 $ \pm $ 0.0027}& \textbf{0.411 $ \pm $ 0.0018} & \textbf{0.2153  $ \pm $ 0.0007} & \textbf{0.49826 $ \pm $ 0.0019}  \\ 
					\midrule			
				\end{tabular}
			}		
		\end{center}
	\end{table*}

\begin{figure*}[!ht]
		\captionsetup[subfigure]{labelformat=empty}
		\subfloat[\footnotesize{(a)}]{%
			\includegraphics[, width=0.25\textwidth]{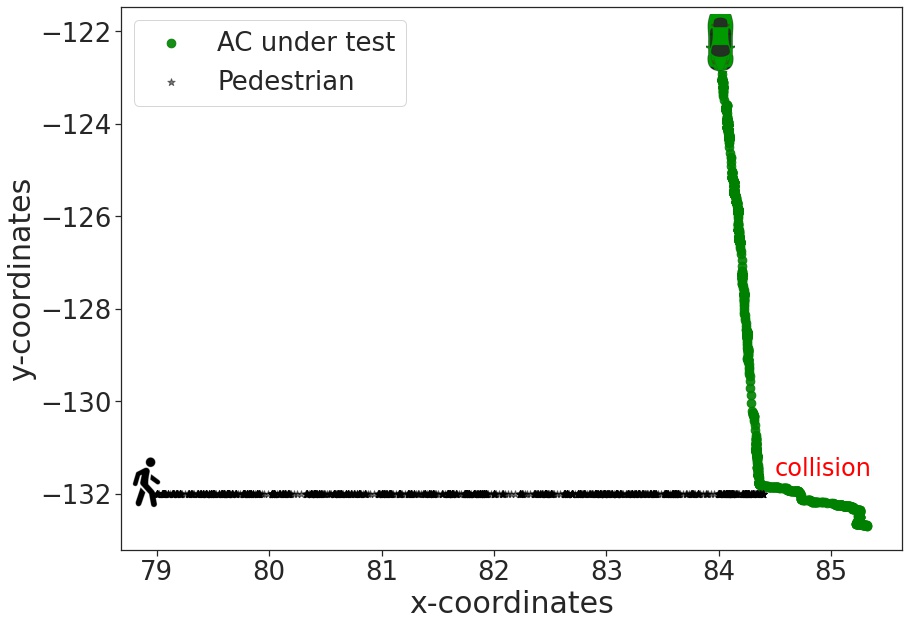}}
		\subfloat[\footnotesize{(b)}]{%
			\includegraphics[width=0.25\textwidth]{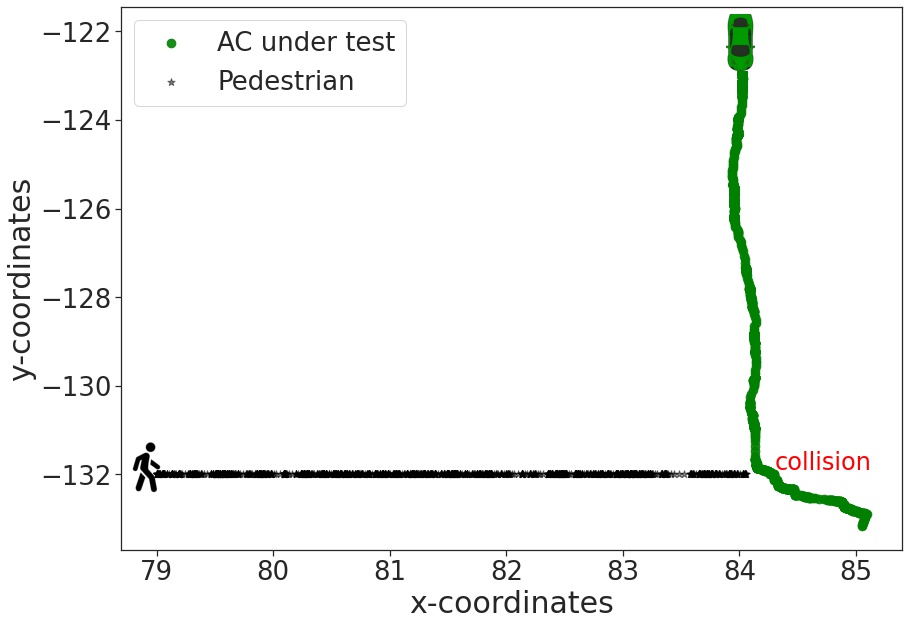}}
		\subfloat[\footnotesize{(c)}]{%
			\includegraphics[width=0.25\textwidth]{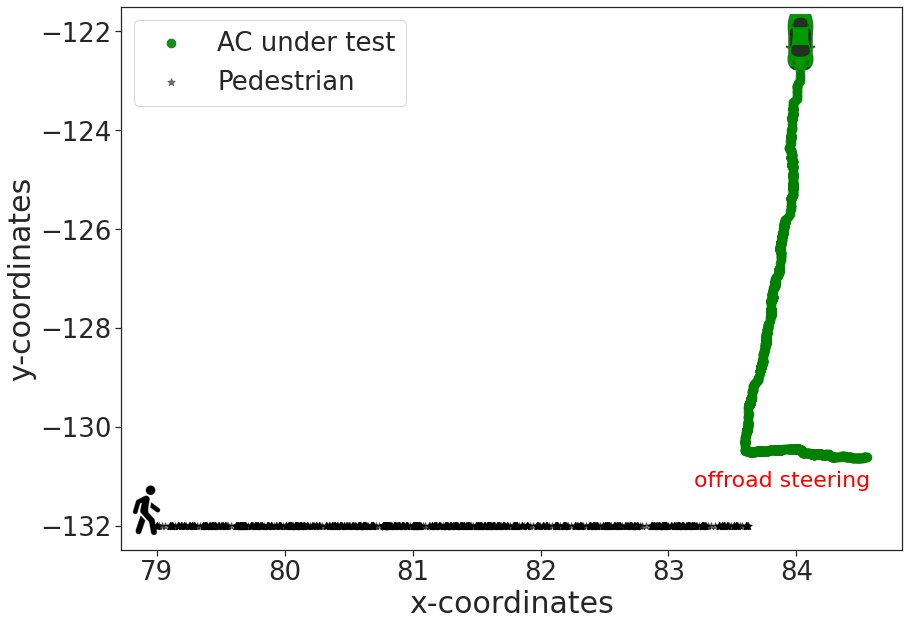}}
		\subfloat[\footnotesize{(d)}]{%
			\includegraphics[width=0.25\textwidth]{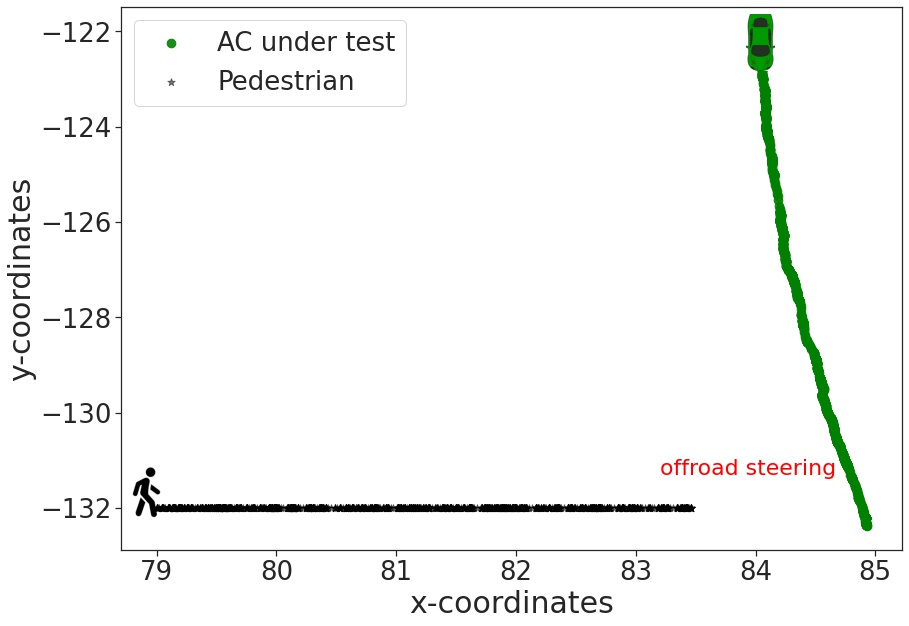}}
		\hspace{\fill}
		\subfloat[\footnotesize{(e)}]{%
			\includegraphics[width=0.25\textwidth]{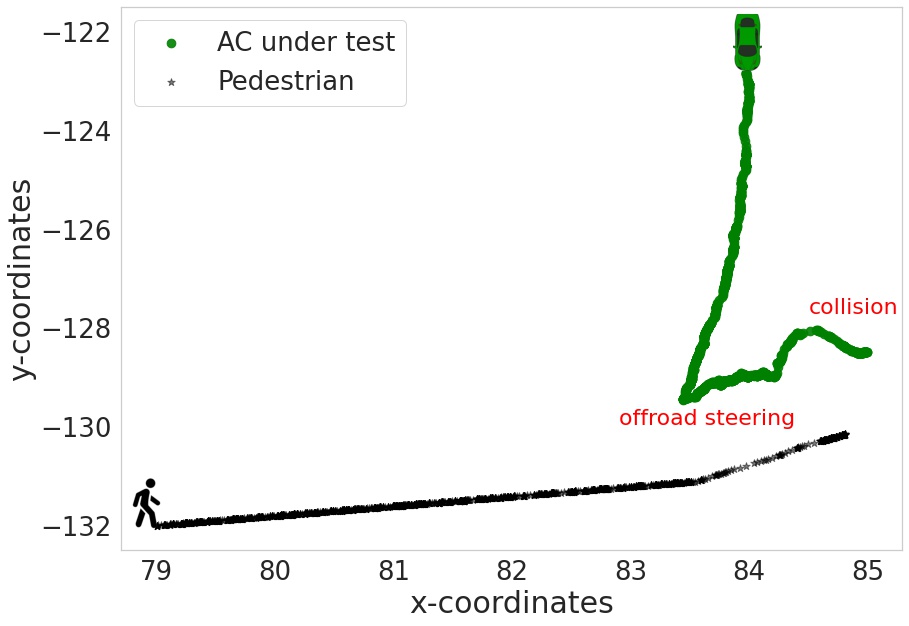}}
		\subfloat[\footnotesize{(f)}]{%
			\includegraphics[width=0.25\textwidth]{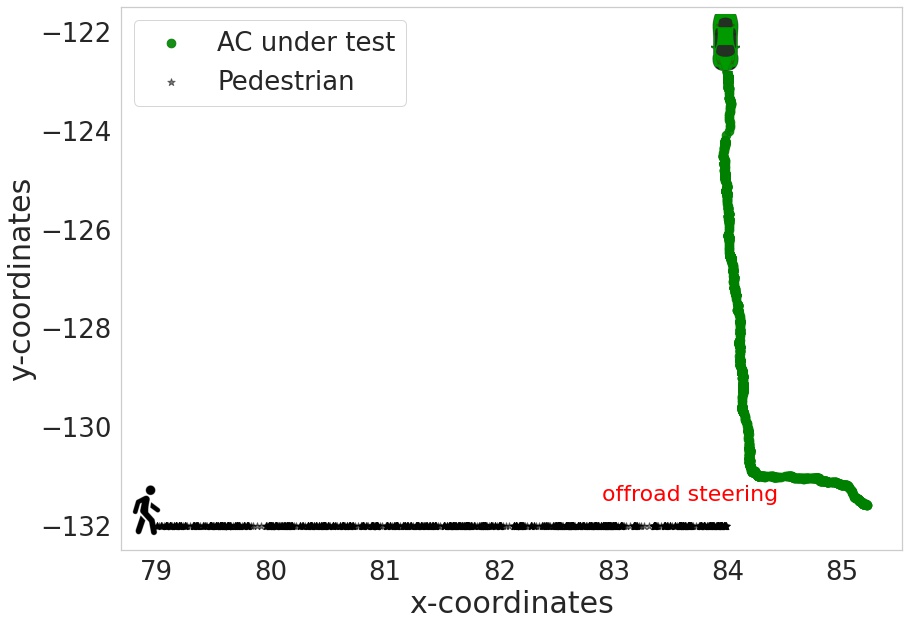}}
		\centering
		\subfloat[\footnotesize{(g)}]{%
			\includegraphics[width=0.25\textwidth]{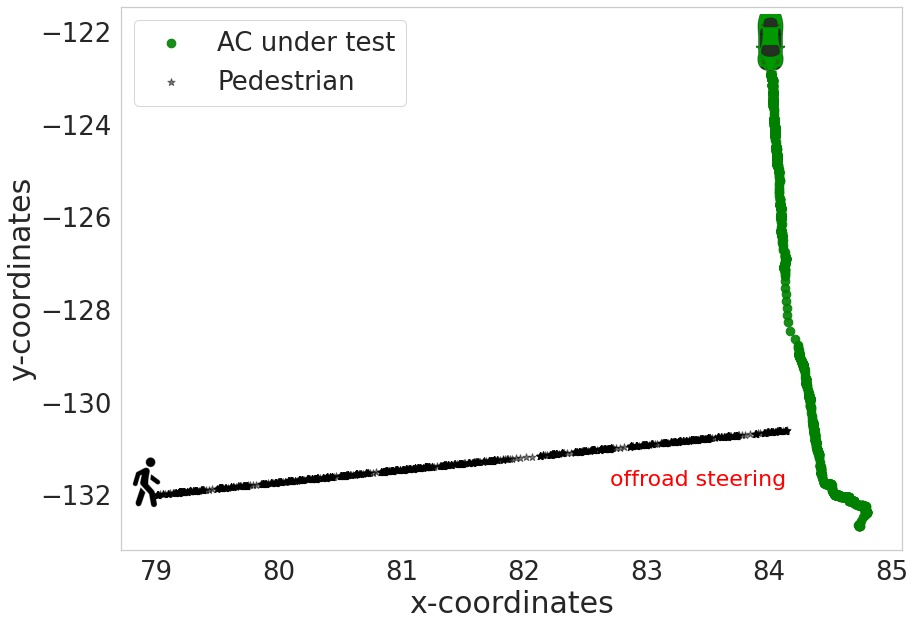}}
		\subfloat[\footnotesize{(h)}]{%
			\includegraphics[width=0.25\textwidth]{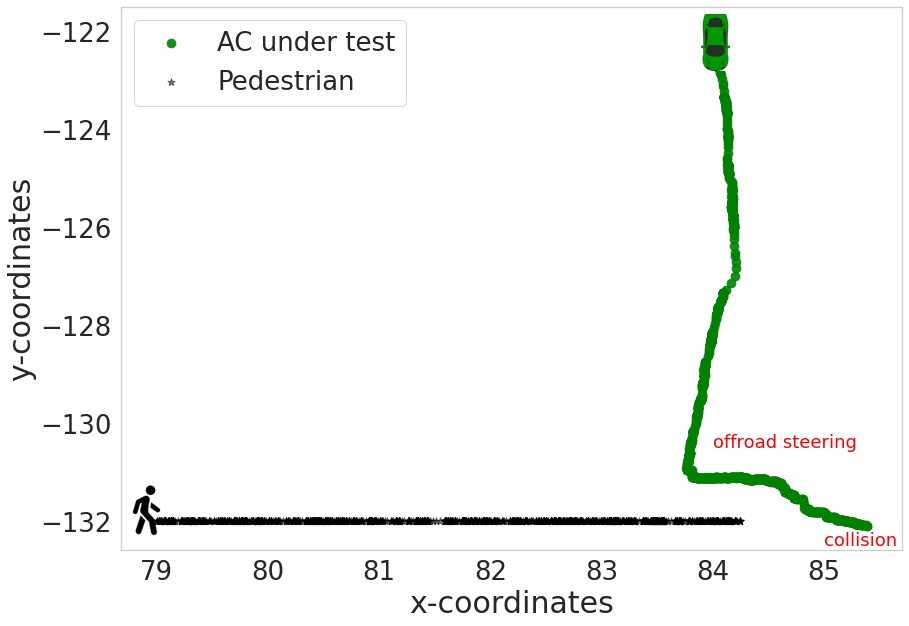}}
		\hspace{\fill}
		\caption{\label{fig:RQ2_3} Top-down visualization of the AV driving coordinates while testing AVs in the presence of disturbance attacks. The testing is performed in a pedestrian driving scenario. (a) and (b) describe the testing episodes in which we observe the collision of the pedestrian with the AV under perturbations of the image noise. To avoid collision, AV steers offroad as shown in (c), (d), (f), and (g). We also find failure events in which the AV collides with road objects while trying to avoid pedestrian collision using extreme offroad steering mistakes as shown in (e) and (h).
		}
	\end{figure*}

	As we move one step towards a multi-agent interaction, we start by first using RT to randomly add noise in the testing episodes of a pedestrian scenario. RT was ineffective when adding image perturbations and was only able to observe rare failure instances. The same can be said when RT was applied for NPC noise perturbations in front of AV under test, where RT again only observed rare failure events. This time with RT we also observed some collisions with road objects. ST performs similarly to RT with few instances of offroad steering errors, pedestrian collisions, and no road object collisions, despite continuous perturbation attacks. Compared to ReMAV simulation testing, the AV under test starts to fail under image and NPC attacks. From the table we see a huge increase in the collision rate in terms of both road and pedestrian. We also see a similar effect regarding the increase in offroad driving events under both noise attacks. In terms of image noise attacks within uncertain states, we observe a 16\% increase in road collisions, while it gets to 23\% road collision events once we also produce challenging scenarios by perturbing the movement of NPC pedestrians within episodes. Furthermore, we show that the average collision rate in all episodes with pedestrians themselves tends to increase by 32 and 48\% under $\epsilon_1$ and $\epsilon_2$ noise. Due to the increase in collisions, in general, we see its side effect in the average increase in offroad steering errors by ~13\%. This can be better explained when visualized in Figure~\ref{fig:RQ2_3} whereas in comparison to Figure~\ref{fig:Baseline} the performance declines, putting safety threads in pedestrian-facing scenarios. Although AV was able to perform well in a straight scenario, we were able to face failure events with a pedestrian scenario using our proposed testing methodology. A common variable in all of the episodes is the lack of braking capability as an action when faced with perturbations in uncertain situations. This led to collisions with pedestrians and offroad steering failures.

\begin{figure*}[!ht]
		\captionsetup[subfigure]{labelformat=empty}
		\subfloat[\footnotesize{(a)}]{%
			\includegraphics[, width=0.25\textwidth]{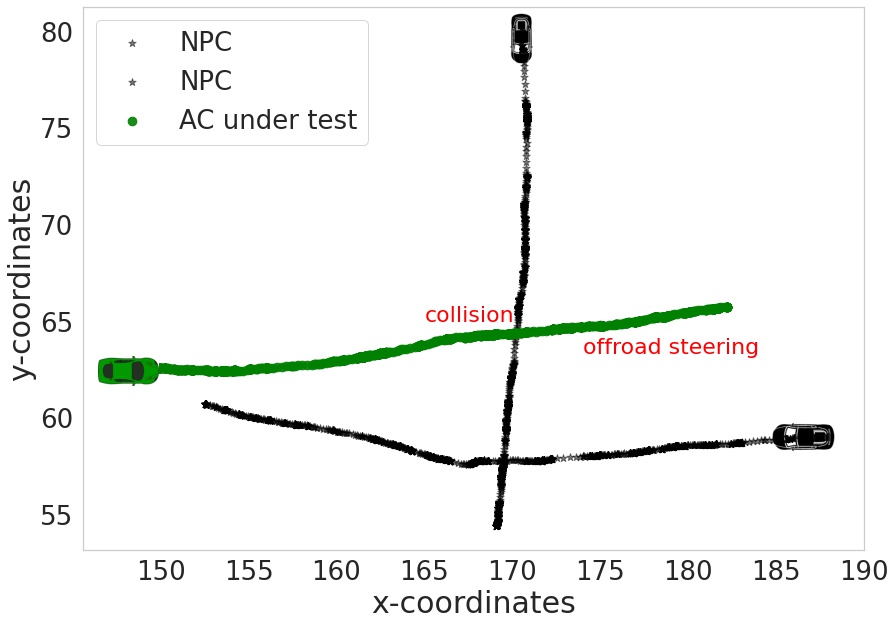}}
		\subfloat[\footnotesize{(b)}]{%
			\includegraphics[width=0.25\textwidth]{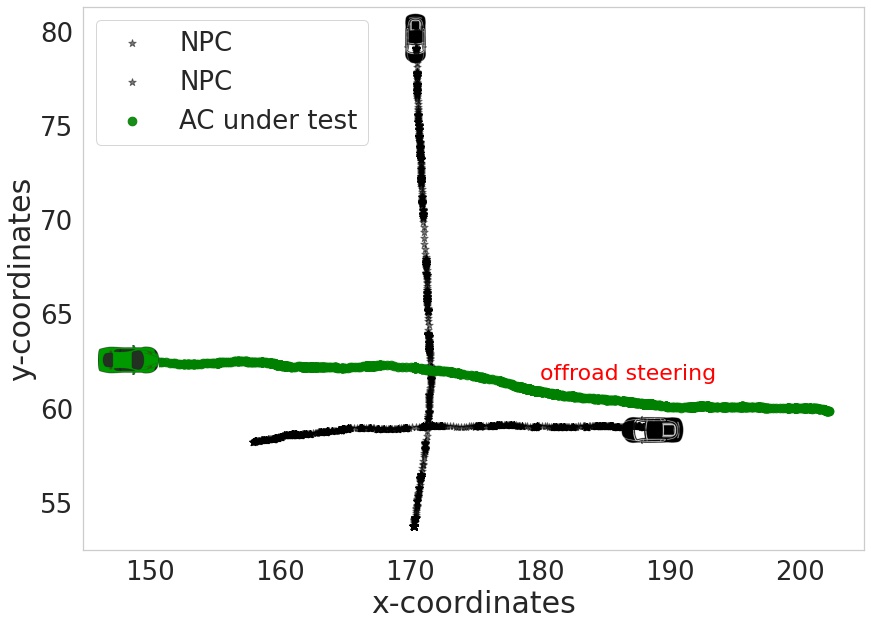}}
		\subfloat[\footnotesize{(c)}]{%
			\includegraphics[width=0.25\textwidth]{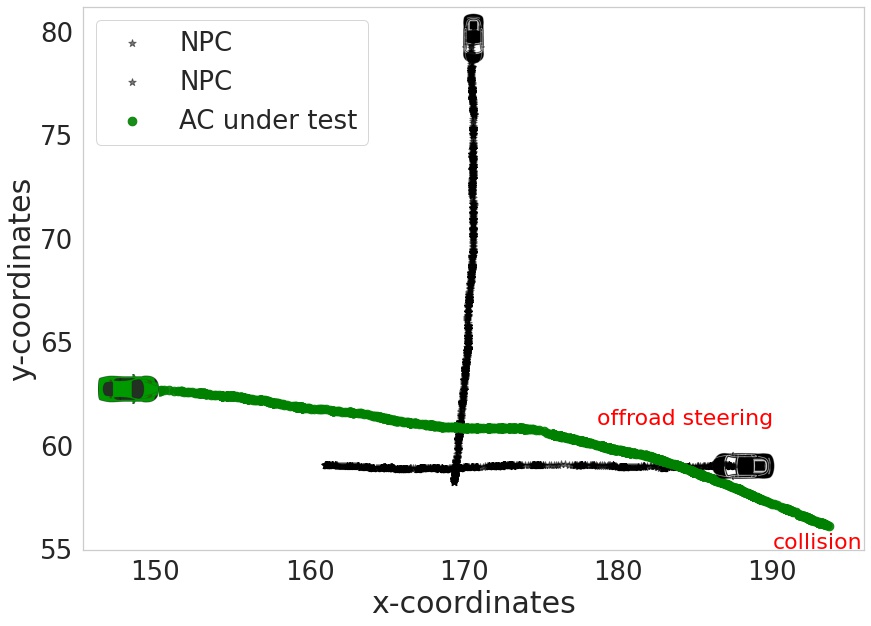}}
		\subfloat[\footnotesize{(d)}]{%
			\includegraphics[width=0.25\textwidth]{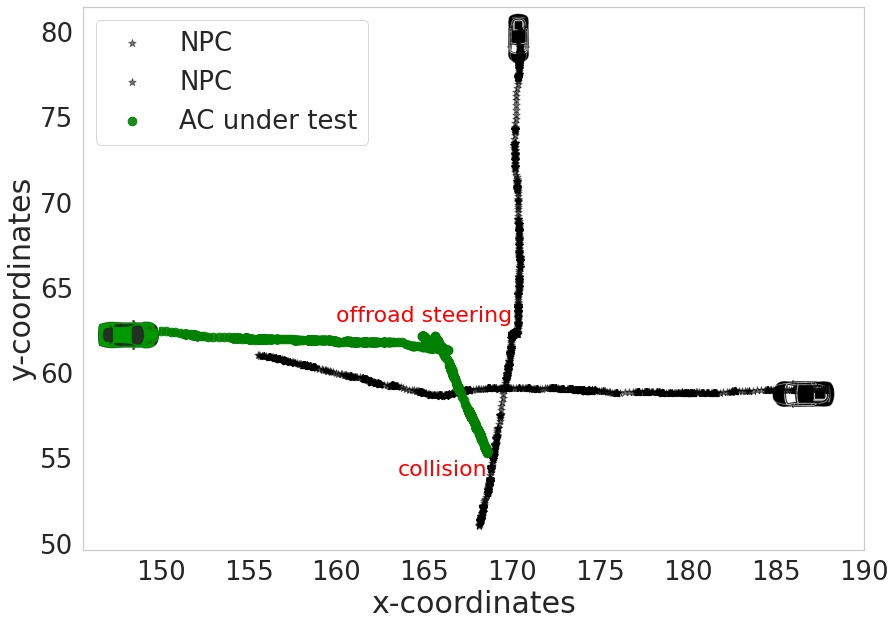}}
		\hspace{\fill}
		\subfloat[\footnotesize{(e)}]{%
			\includegraphics[width=0.25\textwidth]{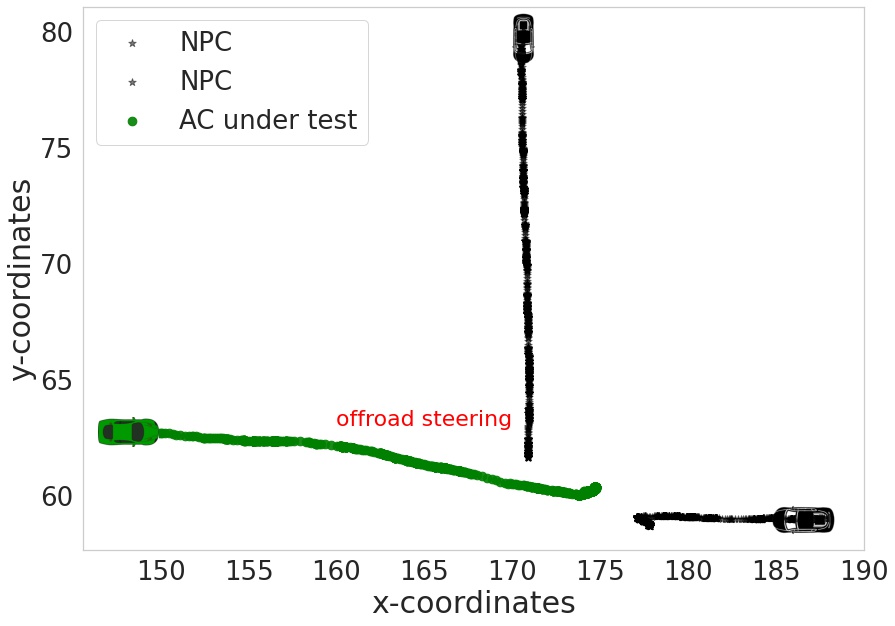}}
		\subfloat[\footnotesize{(f)}]{%
			\includegraphics[width=0.25\textwidth]{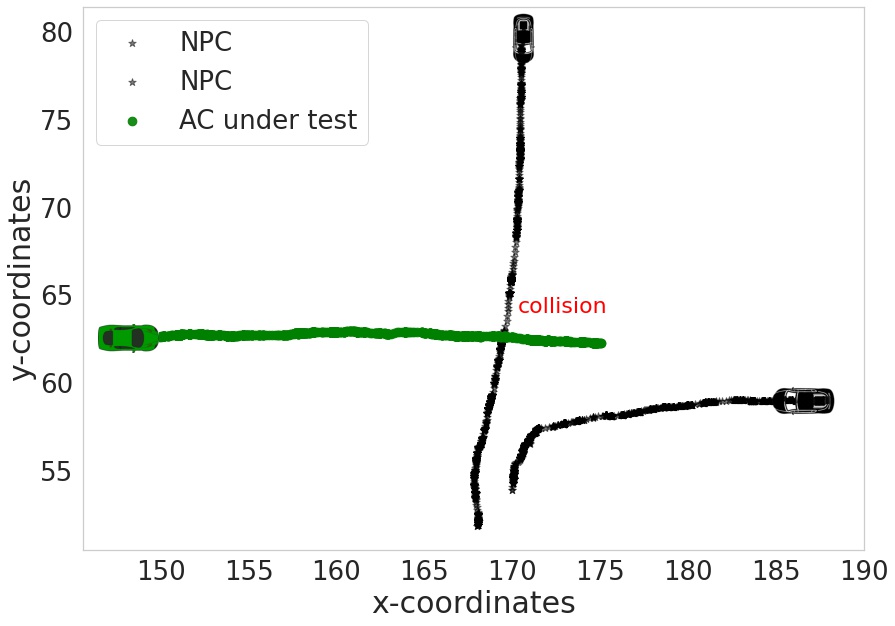}}
		\centering
		\subfloat[\footnotesize{(g)}]{%
			\includegraphics[width=0.25\textwidth]{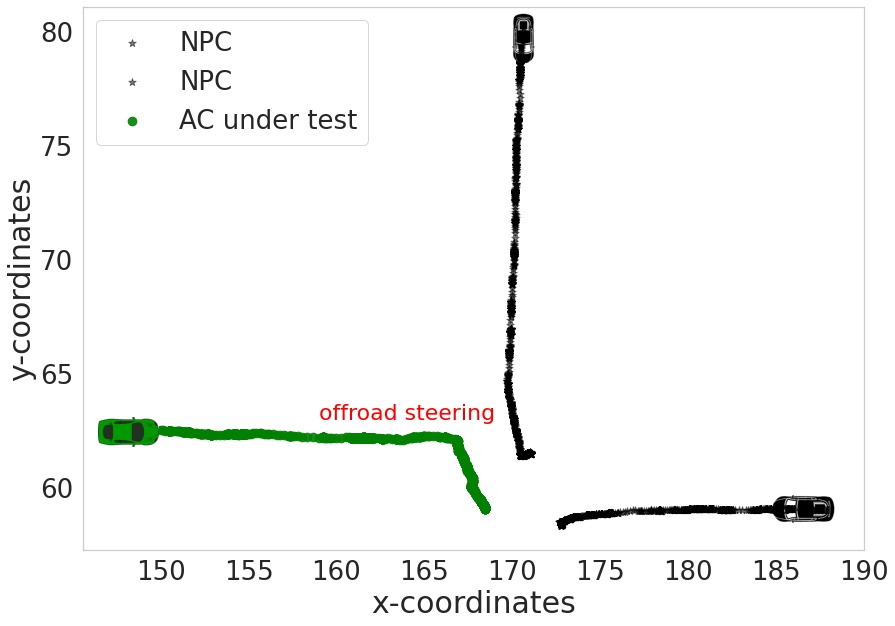}}
		\subfloat[\footnotesize{(h)}]{%
			\includegraphics[width=0.25\textwidth]{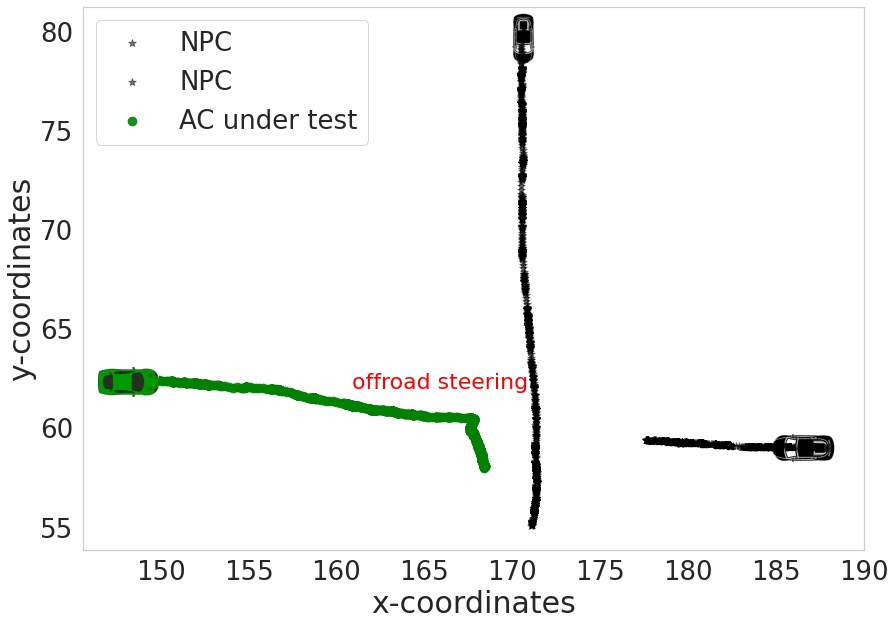}}
		
		\caption{\label{fig:RQ2_4} Top-down visualization of the AV driving coordinates while testing AV in the presence of disturbance attacks. The testing is performed in a three-way driving scenario. We display (f) as the episode with only an NPC collision event and (b), (e), (g), and (h) as the episodes with only offroad steering events. We observe many episodes like (a), (b), (c), and (d) where we face both collision and offroad steering errors when AV faces perturbation attacks.
		}
	\end{figure*}

	We go through the same testing phase for AV in a three-way scenario where more than one NPC driving cars are faced in multi-agent interaction. With RT and ST testing, we found a small number of timesteps in which AV under test went offroad steering and collided with other NPC vehicles. Although RT was able to force AV to crash onto road objects, ST was unable to do the same. As discussed in RQ1, the AV standard driving behavior in three-way was expected to be the weakest among all driving scenarios, but neither RT nor ST were able to find many likely failure events. For ReMAV in comparison, similar to the pedestrian-based scenario, we also experience AV with failure events in the testing episodes. Compared to standard performance, the collision with NPC agents increased to 32 and 35\% when faced with image and NPC agent-based noise attacks. Similarly, we see a significant increase in AV collisions with road objects, with approximately 21 and 17\% occurrences of failure events. Compared to all scenarios, the AV under test faced the highest rate of offroad steering events discovered at an increasing rate of 50 and 30\% for testing under image and NPC noise attacks. The results show that AV in a complex multi-agent environment performed the worst compared to its standard AV performance by a large margin. We see supportive results by visualizing the driving behavior of the AV under test in Figure~\ref{fig:RQ2_4}. As we observed in RQ1, the threshold value deducted from the three-way was an initial indicator of the vulnerability of AV within the three-way scenario. AV performed well as discussed in Figure~\ref{fig:Baseline} but by using the threshold we were able to add minimal noise values to the same AV and it resulted in exposing driving behavior because the agent was least confident in certain states of driving interaction. Figure~\ref{fig:RQ2_4} depicts all types of failure events observed in the testing phase of the framework, especially when the AV was closer to the three-way intersection. As seen in the pedestrian scenario, one of the outcomes of this testing was exposing the AV's incapability to hit brakes under noise perturbation in weak states in most of the episodes. In a three-way scenario, however, we also see that the AV under test was still able to hit the brakes before crossing the intersection point and the agent slowed down but within the episode steps it became confused about the next driving decisions, hence going off-road to stay out of collisions. This is the reason why we see higher offroad steering events in Table~\ref{tab:RQ2_1} compared to all types of collisions.
	
	In summary, we analyze the driving behavior of the AV under test while answering RQ2 in depth. We conclude that the AV under test performs better in single-agent than multi-agent scenarios. We observed a significant rate of collision and offroad instances within the testing phase when AV dealt with noise under uncertain driving conditions and, unlike in the single agent scenario, they were unable to recover to drive well. Lastly, we show that ReMAV performs significantly better in all scenarios compared to baselines in finding collision and offroad steering failure states. We conclude that both types of noise attack from our disturbance model were only effective with ReMAV in generating driving situations to find the most likely failure events.

	\subsection{RQ3: Does the reward function help reduce the search space for finding edge cases compared to baseline approaches?}

	Once we have analyzed $R_\psi$ and the required thresholds, we can answer RQ3 using Table~\ref{tab:RQ3_1}. The table makes a comparison between two components; i) the time in seconds it takes to add the first attack, and ii) the average time in seconds to observe the first failure event.
	
	As described in Section~\ref{sec:Step3}, by using $R_\psi$ and $\beta$ as thresholds per scenario, we simulated AV under test and searched for state-action pairs that provide a reward value below the desired threshold. For the straight driving scenario, which is a single agent driving situation with no NPCs and pedestrians involved, we only attack using the image disturbance model with minimal perturbation noise $\epsilon_1$ sampled from the Gaussian distribution. For the pedestrian scenario, the AV under test is facing a pedestrian crossing a road and therefore we use not only $\epsilon_1$ but also $\epsilon_2$ as the minimal noise perturbed within the actions of the pedestrian as NPC. Similarly to this, we use $\epsilon_1$ and $\epsilon_2$ for noise attacks on the AV under test in the three-way scenario.

		
	Since RT attacks by selecting random states per timestep, it usually adds the first image or NPC actions noise within 2 seconds of the testing episodes. Similarly, ST stress tests each state space of timestep in the testing phase and therefore starts adding the first noise in less than a second. As for ReMAV, in terms of seconds, we see that the AV in the straight scenario was mostly less confident at the start of the episodes, therefore, the noise has been added by ReMAV earlier while $R_\psi$ crossed the threshold $\beta$ negatively. However, in the pedestrian scenario, it takes $\approx$ 128 seconds for the image noise and $\approx$ 55 seconds for NPC action noise to be added to the simulation testing. Whereas in the three-way, the testing phase takes $\approx$ 27 and $\approx$ 36 seconds for each noise perturbation on average to attack the AV under test.
	
	To measure the robustness of AV according to their driving scene, we look at the bottom results of Table~\ref{tab:RQ3_1} for a clear picture. AV under test appears to perform well in a single-agent straight scenario where it does not collide with any road objects. Instead, we see offroad steering errors that occur primarily in the starting and ending phases of the testing episodes, which were only found during the ReMAV testing phase.
	
	A single episode while testing AVs takes around 200 seconds. As we go one level above for testing AV driving complexity by adding a pedestrian, we do observe AV under ReMAV colliding with pedestrians and road objects in the testing episodes. Near the end of the episodes, we also see an offroad steering error being performed by AV under test. RT on the other hand takes around the end of the episode to detect the first collision and also takes multiple episodes to observe the first offroad steering error state. ST performs worse in comparison by taking multiple episodes to observe the first collision and more than 30 minutes of the testing phase to detect the first offroad steering error.
	
	Finally, we evaluate the robustness of AVs by going a level up as in three-way driving. From the table, we see that AVs under ReMAV testing go into their first collision in the first quarter of the episode run, while they also face first offroad steering errors while reaching half of the episodic runs. RT takes $\approx$ 45 and 33 minutes of the testing episodes to observe the first collision and offroad steering state respectively. ST takes similarly $\approx$ 24 and 60 minutes to detect the first failure scenario.
	
	In summary, Table~\ref{tab:RQ3_1} conveys that AVs will be attacked at different times on average under different driving conditions based on their least confident interactions during testing. We also show that ReMAV compared to both baselines takes a shorter time to observe the first failure states.
	
	\begin{table*}[!htbp]
		
		\begin{center}
			\caption{Comparison of the behavior of AVs under test in three scenarios. The top row shows the average time in seconds it takes to add the first attack. The two bottom rows in the table represent the time in seconds to observe the first failure event. The double dashed (- -) shows the testing phase in which the failure event was not observed by the testing techniques. } \label{tab:RQ3_1}
			\resizebox{!}{3.4cm}{
				\begin{tabular}{ll|l|l|l}
					
				&	& \multicolumn{1}{c}{Straight}   & \multicolumn{1}{|c}{Pedestrian} & \multicolumn{1}{|c}{Three-Way} \\
					

					\midrule
			\multirow{2}{*}{ReMAV}	&	Average Time to add fist image noise (seconds) & 2.01 & 128.64 &  27.604  \\
				&	Average Time to add the first NPC noise (seconds) & - &  55.476   &  36.716   \\
				\cmidrule(lr{1em}){2-2}
			\multirow{2}{*}{RT}	&	Average Time to add fist image noise (seconds) & 1.05 & 2.0 & 2.05   \\
				&	Average Time to add fist NPC noise (seconds) & - &  1.0   &  2.5  \\				
				\cmidrule(lr{1em}){2-2}
			\multirow{2}{*}{ST}	&	Average Time to add fist image noise (seconds) & 0.5 & 0.5 & 0.5   \\
				&	Average Time to add the first NPC noise (seconds) & - & 0.5 & 0.5    \\
					\midrule
			\multirow{2}{*}{ReMAV}	& TTFC (seconds) & - - & \textbf{132.258} &  \textbf{47.436}    \\
				& TTFO (seconds) & \textbf{16.08} & \textbf{176.21} &  \textbf{82.812} \\
				\cmidrule(lr{1em}){2-2}
			\multirow{2}{*}{RT}	& TTFC (seconds) & - - & 180.5 & 2706.5     \\

				& TTFO (seconds) & - - & 1830.0 & 1997.0   \\					
				\cmidrule(lr{1em}){2-2}
			\multirow{2}{*}{ST}	& TTFC (seconds) & - - & 428.19 &  1483.43    \\
				& TTFO (seconds) & - - & 658.141 & 3606.744  \\
				\bottomrule
				\end{tabular}
			}		
		
		\end{center}
	\end{table*}
	
\begin{figure}[!b]
		\centering
		
		\includegraphics[width = 0.5\textwidth,height=0.26\textheight]{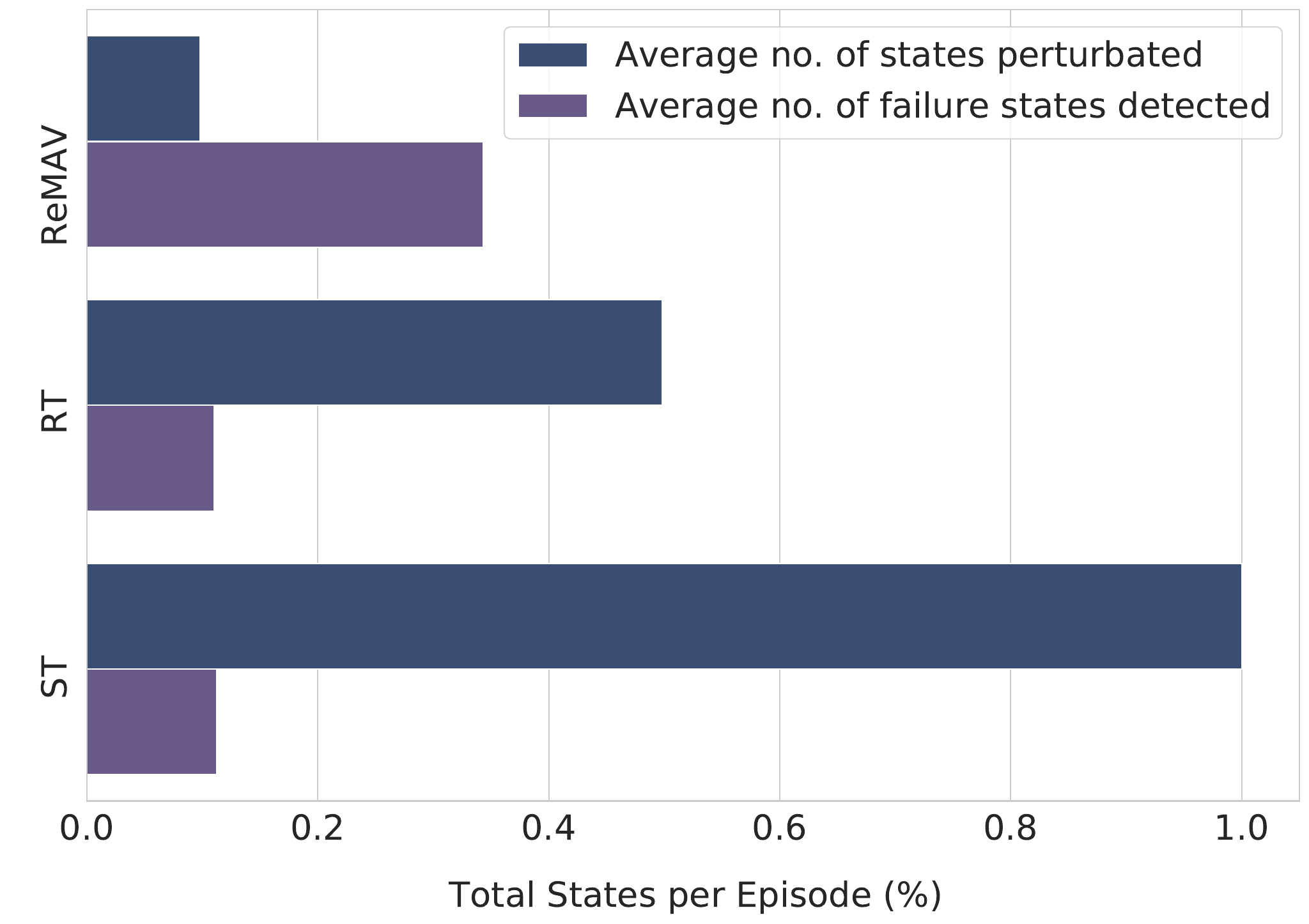}
		\caption{Comparison of the behavior of AVs under test in three scenarios. The top bar against each testing technique defines the amount of noise perturbed within all three driving scenarios. The bottom bar for the comparison technique shows the average number of likely failures detected.}
		\label{fig:barplot}
	\end{figure}

	We can also visualize the average amount of noise on average (both in image and NPC action) perturbated within all three driving scenarios versus the average number of failure states discovered. According to Figure~\ref{fig:barplot}, ReMAV performs much better by attacking only 9.8\% of the states during the simulation testing phase. In return, it detects 34.3\% failure events. Compared to baselines, RT and ST on average attack 49.8\% and 100\% of the testing phase whereas they only detect 11\% and 11. 2\% failure driving states, respectively.

%
%
%
%

\begin{figure}[!htp]
		\captionsetup[subfigure]{labelformat=empty}
		
		\subfloat[\footnotesize{(a)}]{%
			\includegraphics[width=0.166\textwidth]{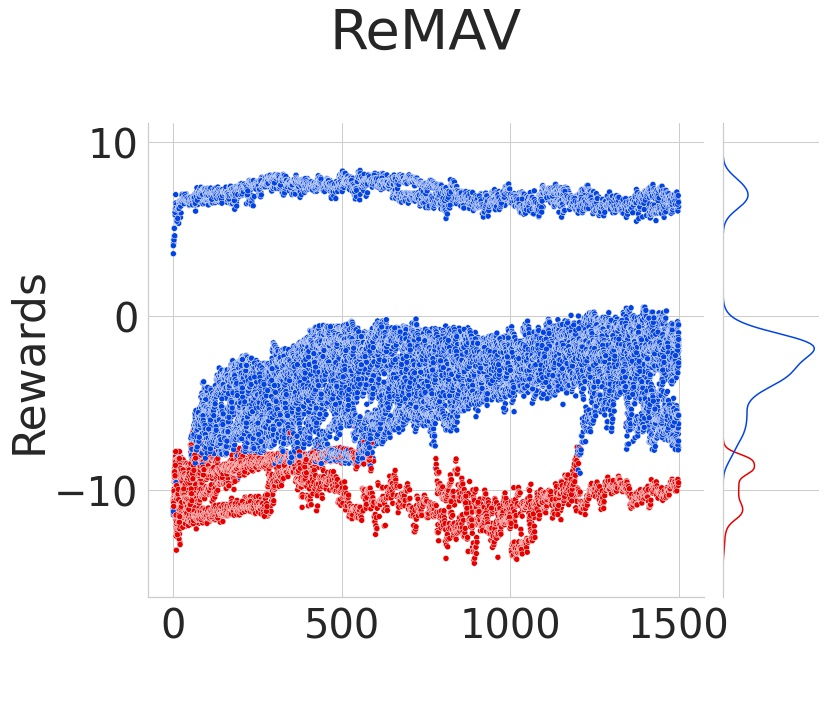}}
		\subfloat[\footnotesize{(b)}]{%
			\includegraphics[width=0.166\textwidth]{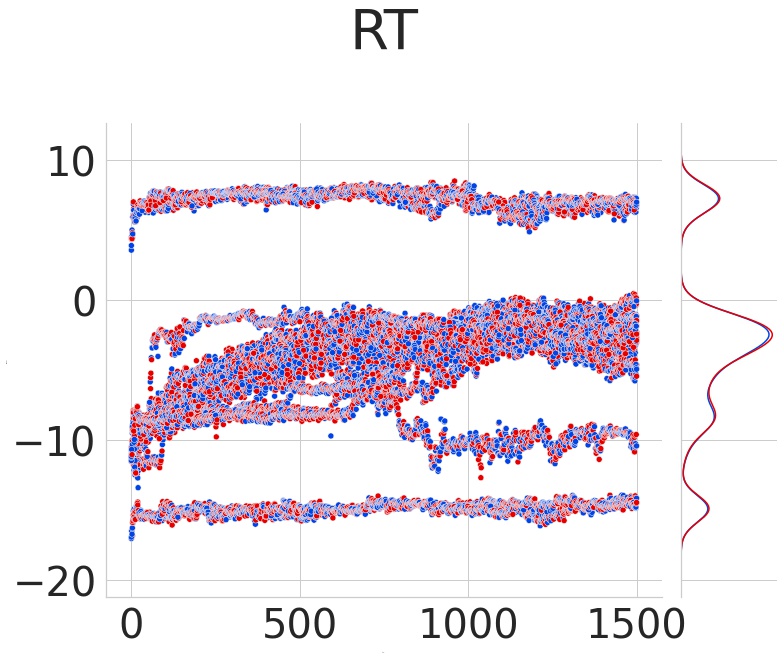}}\subfloat[\footnotesize{(c)}]{%
			\includegraphics[width=0.166\textwidth]{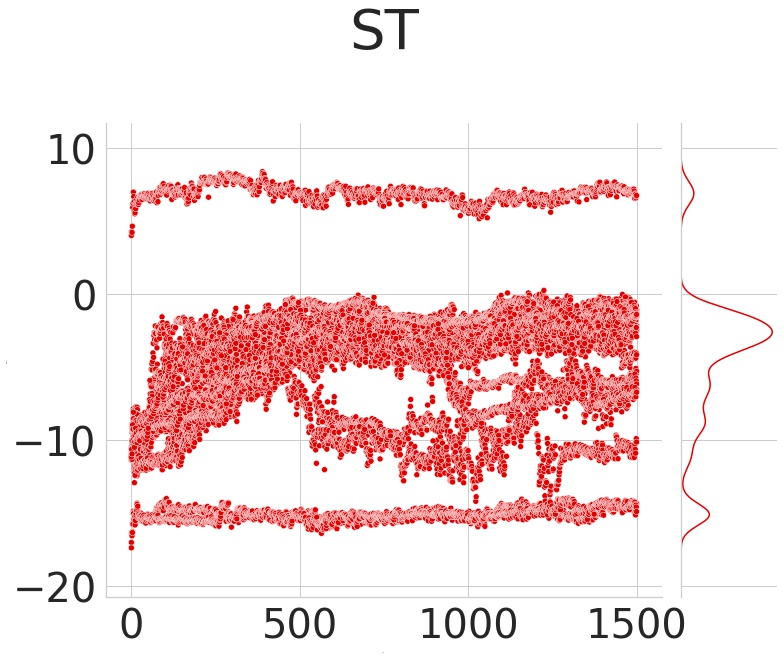}}
		\hspace{\fill}
		\centering
		\text{\small{Straight}}

				\captionsetup[subfigure]{labelformat=empty}
		\subfloat[\footnotesize{(d)}]{%
		\includegraphics[ width=0.166\textwidth]{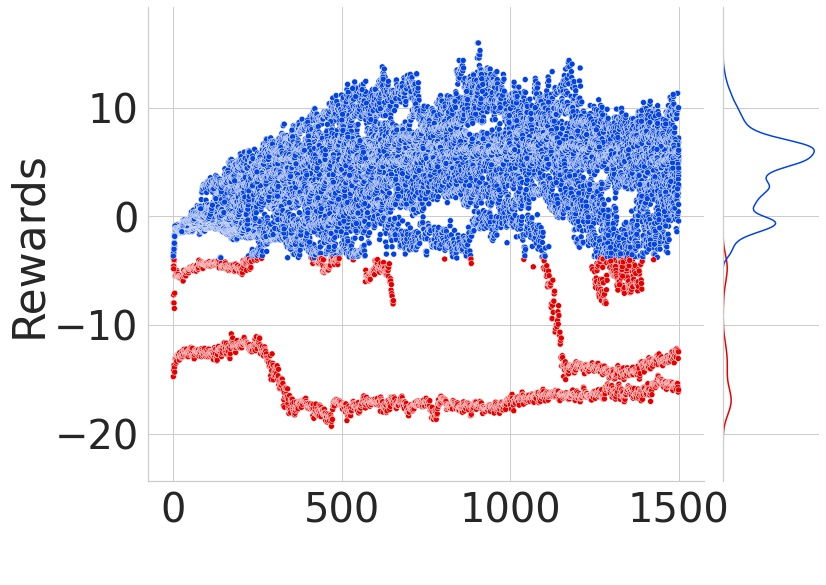}}
		\subfloat[\footnotesize{(e)}]{%
			\includegraphics[width=0.166\textwidth]{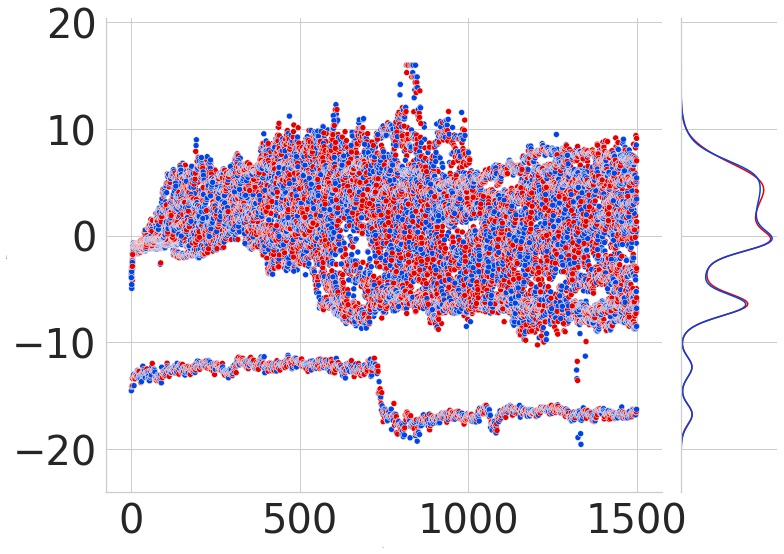}}\subfloat[\footnotesize{(f)}]{%
			\includegraphics[width=0.166\textwidth]{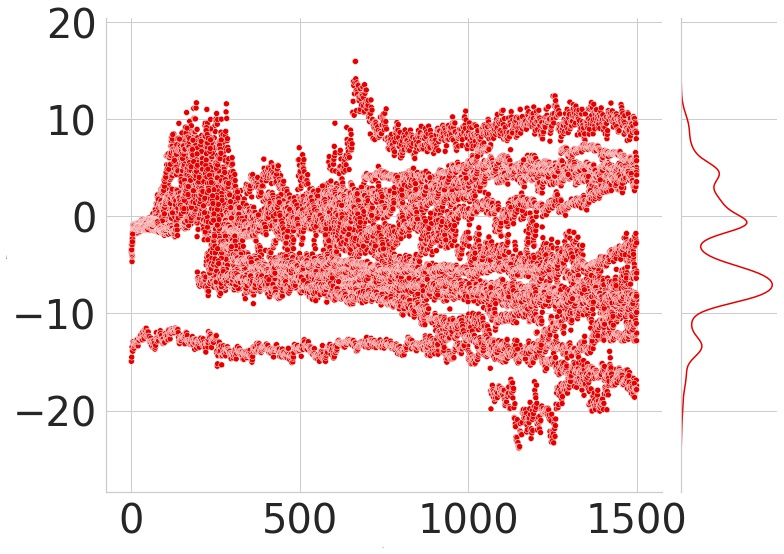}}
		\hspace{\fill}
		\centering
		\text{\small{Pedestrian}}
		
						\captionsetup[subfigure]{labelformat=empty}
		\subfloat[\footnotesize{(g)}]{%
			\includegraphics[ width=0.166\textwidth]{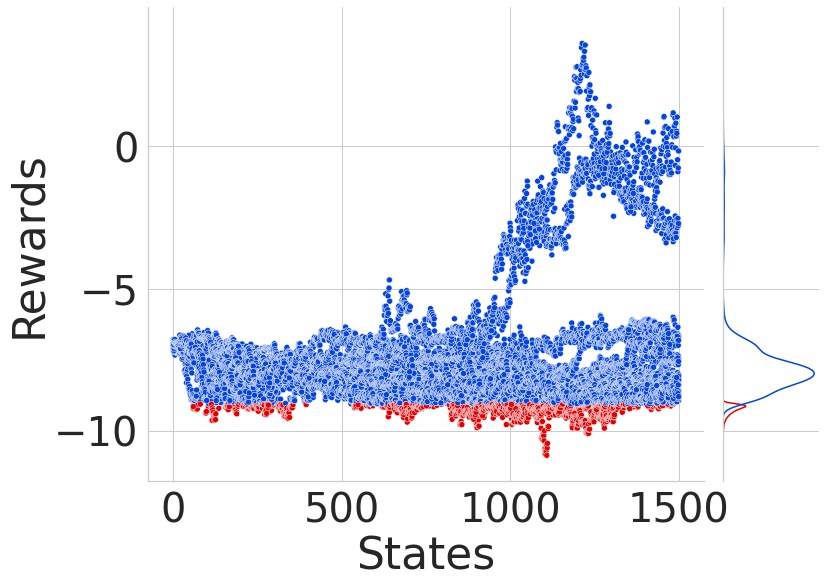}}
		\subfloat[\footnotesize{(h)}]{%
			\includegraphics[width=0.166\textwidth]{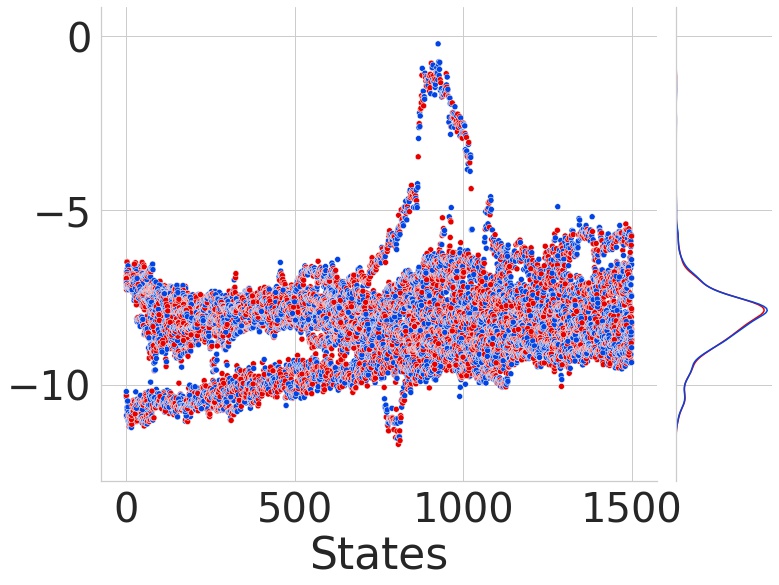}}\subfloat[\footnotesize{(i)}]{%
			\includegraphics[width=0.166\textwidth]{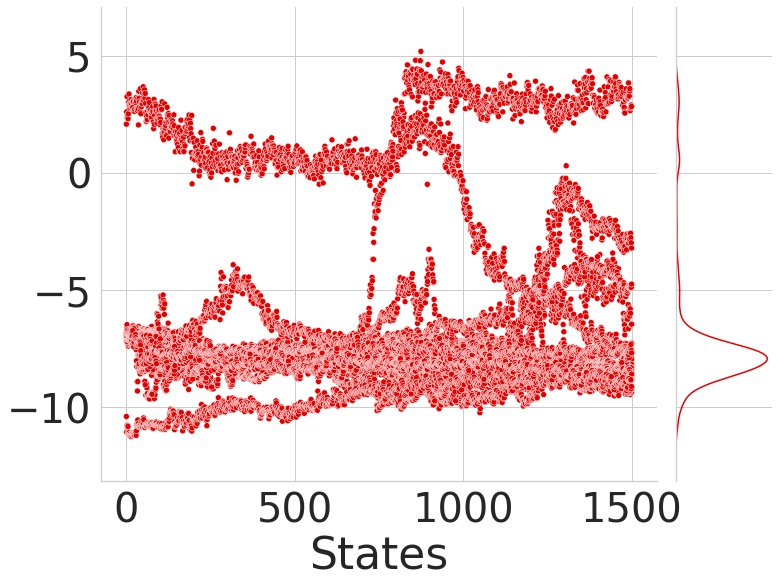}}
		\hspace{\fill}
		\centering
		\text{\small{Three-Way}}
		\caption{\label{fig:RQ3_2} Visualizations of $R_{\psi}$ values with respect to the noise added in the respective state space using ReMAV, RT, and ST. For all three rows as scenarios, red dots are the ones where noise is added using our disturbance model. (a), (d), and (g) shows the least amount of noise attacks by ReMAV during the testing phase. 
		}
	\end{figure}

	We also address RQ3 by showing the noise distribution within the testing phase in Figure~\ref{fig:RQ3_2}. Each row corresponds to the three scenarios that we have used in our experiments. The figures in each row display the reward values with the states visited. The red dots in the figures highlight the state in which the noise was added. The figures on the left show the simulation tests performed by ReMAV where the blue dots are normal states with $R_\psi$ values above their respective thresholds. We see that in the straight scenario, the noise was added the most compared to the other two scenarios, but the AV under test performed well by showing resilience to such attacks. Whereas in multi-agent scenarios as discussed in Figure~\ref{fig:barplot} very low amount of noise has been added in the testing phase. By first detecting uncertain areas of the AV under test, we were able to highlight more failure events even with such a lower amount of strategic noise perturbations. Inversely, RT in the middle figures targets almost half of the states, whereas ST adds noise to each state in an episode.

	In short, we can visualize the reduction in the search space for testing using a simplistic noise disturbance model as in our framework to find failure states. As discussed in our methodology, a key contribution of our work is to minimize the number of adversarial attacks to find any failure event by first identifying the existing flaws of the AV under test. The amount of noise required for the validation of the robustness of AV after ReMAV identifies the states under uncertainty is reduced a lot. As extensively shown in Table~\ref{tab:RQ3_1} and Figure~\ref{fig:barplot}, ReMAV performs better than the baseline methods when it comes to finding the first failure state as well as finding the most likely failure events with the least amount of noise required.

	\subsection{How does ReMAV perform compared to the existing AV testing frameworks?}\label{sec:RQ4}

	Here we delve into the comparative analysis of ReMAV with two existing works. 
	
	\subsubsection{Training and testing efficiency}~\label{firstcomparison}
	
	First, we compare the computational efficiency of training and testing between our proposed method, ReMAV, and BayesOpt. Table~\ref{tab:RQ4_1} shows a comparison of the iterations required by ReMAV and BayesOpt to train and evaluate AVs in a straight scenario. We see that BayesOpt needs 400 iterations to learn to attack AVs in a straight-driving scenario, while ReMAV needs $\approx$166 training iterations to capture the AV's straight-scenario behavior. We also see that ReMAV takes a total of 500 training iterations to learn three driving scenarios before attacking AVs. ReMAV's total testing phase is performed using 150 iterations with 50 iteration episodes per scenario. This shows that, without actively looking for adversarial attacks while training the algorithm, ReMAV first uses offline data to understand the driving behavior of the AV in each scenario.

	\begin{table}[!htbp]
		
		\begin{center}
			\caption{Comparison of the training and testing computational efficiency of ReMAV against BayesOpt. The comparison is carried out within a straight driving scenario.} \label{tab:RQ4_1}
			\resizebox{!}{0.95cm}{
				\begin{tabular}{l|l|l}
					
					& \multicolumn{1}{c}{Training}   & \multicolumn{1}{|c}{Testing}  \\
					
					\midrule
					BayesOpt (iterations)  & 400 & -   \\
					ReMAV  (iterations) & 166 & 50   \\
					\midrule

				\end{tabular}
			}		
		\end{center}
	\end{table}

	\subsubsection{Total infractions detected}
	
	We compare ReMAV and BayesOpt using a three-way intersection scenario to evaluate the number of infractions detected by each algorithm. Table~\ref{tab:RQ4_2} compares these algorithms using CO and OS as evaluation metrics. We observe significant differences in detecting collision and offroad steering errors when comparing them side by side.
	
	\begin{table}[htbp]
		\caption{Comparison of ReMAV and BayesOpt with respect to the total number of infraction scenarios detected during three-way intersection scenario.} \label{tab:RQ4_2} 
		
		\begin{center}

			\resizebox{!}{1.6cm}{

				\begin{tabular}{ll}
					
					\toprule
					
					\textbf{} & \textbf{\# of infractions} \\ \midrule

					\multirow{2}{*}{BayesOpt} & 2.5\% CO in 600 episodes  \\ 
					
					& 2.5\% OS in 600 episodes    \\

					\midrule \midrule
					
					\multirow{2}{*}{ReMAV}  &	24.9\% CO in 50 episodes  \\ 
					
					& 69.8\% OS in 50 episodes   \\

					\bottomrule
					
				\end{tabular}

			}

		\end{center}

	\end{table}
	
	BayesOpt performs 600 testing iterations to observe 2.5\% of the infractions of the AV under test. On the contrary, ReMAV detects 24. 9\% of collisions with objects and 69.8\% of offroad steering events in just 50 testing iterations using image noise attacks $\epsilon_1$. Therefore, we demonstrate that by first modeling the reward representation of the AV under test, we can identify uncertain states and maximize the chances of finding failure events using image noise attacks.

	\subsubsection{Simulation steps to find first failure event}
	
	Finally, we compare our results with the AST-BA algorithm. The goal of both ReMAV and AST-BA is to find events in which the AV under test fails to perform normal driving. AST-BA adds noise to the movement of the pedestrian NPC, as also done by ReMAV using $\epsilon_2$. As described in~\ref{sec:ReMAV}, ReMAV learns the reward model of the AV first and then analyzes the distributions to plan strategic attacks, unlike many testing optimization algorithms. 
	
\begin{figure}[!t]
		\captionsetup[subfigure]{labelformat=empty}
		\subfloat[\footnotesize{(a)}]{%
			\includegraphics[, width=0.24\textwidth]{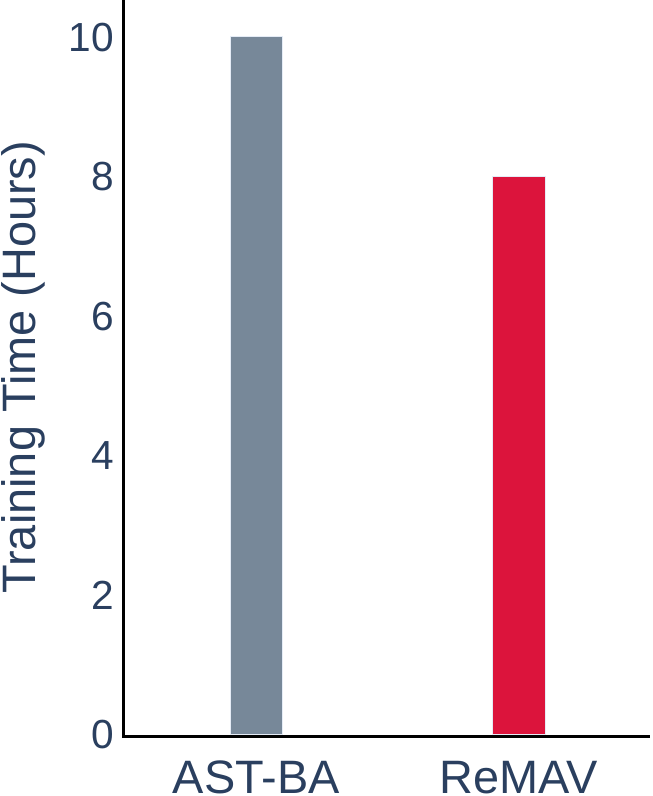}}
		\vspace{\fill} \vspace{\fill}
		\subfloat[\footnotesize{(b)}]{%
			\includegraphics[width=0.24\textwidth]{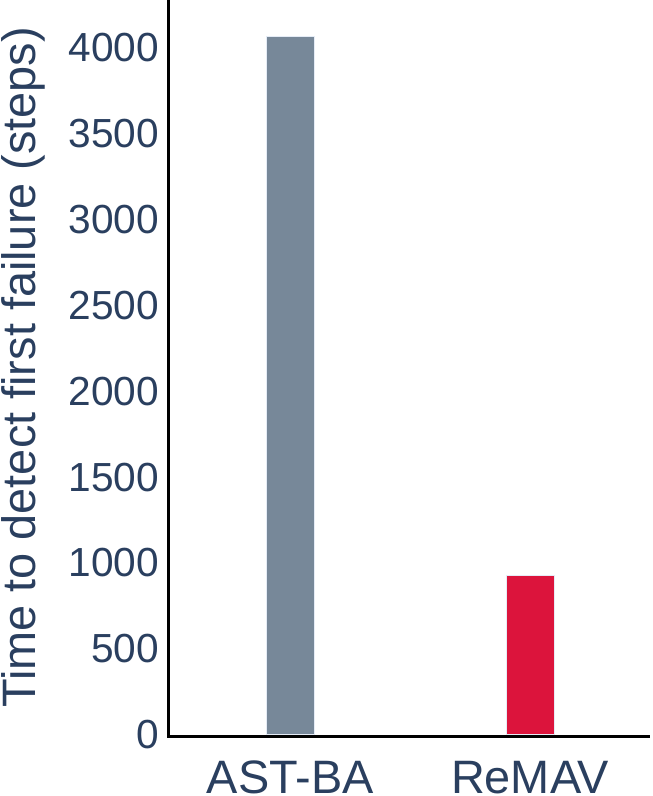}}
		\hspace{\fill}
		
		\caption{\label{fig:RQ4_3} Comparison of the ReMAV and AST-BA testing frameworks in terms of training hours and the time to detect the first pedestrian AV collision.}
	\end{figure}

	In Figure~\ref{fig:RQ4_3}, we represent the time it takes to find the first failure event in a pedestrian driving scenario. AST-BA takes up to 10 hours of training to find the first failure event. On the other hand, ReMAV's overall training takes up to 8 hours, where $\approx$2.6 hours are captured by the offline trajectories of pedestrian scenarios. We also show that in terms of simulation timesteps, AST-BA reports 4060 steps to observe the first pedestrian collision with an AV under test after exhaustive training. ReMAV identifies the first collision event in 925 simulation steps (approximately 55 seconds) during its testing phase. On the basis of this comparison, we show that ReMAV performs better in attacking AV within complex driving situations such as pedestrian crossings.

	In conclusion, ReMAV's efficiency in training and testing is compared to BayesOpt. ReMAV is found to require significantly fewer iterations to capture the autonomous vehicle's driving behavior. Additionally, when subjected to image noise attacks in a three-way intersection scenario, ReMAV detects more infractions than BayesOpt. Finally, ReMAV is compared to the AST-BA algorithm, in which it is found to perform better in attacking AVs within complex driving situations such as pedestrian crossings. Overall, ReMAV is demonstrated to be an effective method for attacking autonomous vehicles.

	Despite having comparisons discussed above, it is important to mention that both existing frameworks cannot be equally compared due to the following reasons:
	\begin{enumerate}
		\item Proposing a novel testing framework as a prior step to existing adversarial techniques has not been done before, limiting any direct comparison.
		\item Each methodology depends on different sets of objective goals for the training and evaluation of the proposed testing frameworks.
	\end{enumerate} 
	Instead, we compare them based on common grounds available, such as driving scenarios for experimentation, metrics used for evaluation, and training-testing performance.


	\section{Threats to Validity}
	
	This section discusses potential threats to the validity of the findings presented in this study.

	\subsection{Internal} We might have introduced errors in the implementation of the ReMAV methodology, from collecting offline trajectories to reward modeling of AV under test as well as the simulation testing phase. To mitigate this threat, we used the widely-used TensorFlow and Rllib frameworks within the Carla simulator for algorithm development and carefully examined the implementation. Furthermore, the implementation of ReMAV~\footnote{https://github.com/T3AS/ReMAV} will be publicly available for inspection and use as an open source project after acceptance on paper. The second threat to validity is the quality of the offline trajectory dataset used in ReMAV training. We collected offline trajectories of state-action pairs within the driving simulator where the AV has been trained. To mitigate the threat of bias in the ReMAV training phase, we performed our experiments in three different driving scenarios for added diversity. Third, hyperparameter tuning plays an important role in achieving good model performance. In our work, as mentioned in Section~\ref{sec:ReMAVExperimentalSteps}, we describe hyperparameters in depth, from the AV standard behavior training to the ReMAV algorithm training and simulation testing setup. Using different hyperparameters may lead to different results. However, we experimented with many different configurations while training ReMAV and selected the best-performing set of hyperparameters.
	
	\subsection{External} Threats to external validity relate to the generalization of the proposed ReMAV framework for testing AVs. Our study focused on specific driving scenarios based on tackling varied complexity (from single-agent to multi-agent with NPC cars and pedestrians) and diversity in scenarios (straight road, T-intersection), which might not fully represent the range of situations encountered in overall driving situations. Therefore, the performance of ReMAV observed in our study might not necessarily reflect its performance in all possible scenarios. Moreover, our work requires the testing of image-based AVs in a multi-agent simulation environment. Such a software architecture has been recently proposed, and thus limits the usage of ReMAV for directly importing existing AVs trained in the Carla simulator by the open-source community. However, our proposed work is designed in a way where the methodology is independent of the simulator, and thus making ReMAV more modular to be used for AV testing is planned for future work.
	
	\subsection{Construct} For our experiments, we used a DRL-based AV policy under test. As an extension, we aim to target other types of AV architecture under test to further validate our approach. Furthermore, we used six evaluation metrics to gather our results. In different environments and case studies, a different set of evaluation metrics and statistical analysis could be relevant. We will consider both threats as a future direction.
	
	\subsection{Conclusion}\label{sec:Conclusion}
	Despite our comparison to different testing techniques as discussed in Section~\ref{sec:RQ4}, our work is still bound by certain technical limitations that present significant challenges when making direct comparisons to the rest of the published ideas. These limitations encompass factors beyond our control, such as the difference in experimental setup for training and evaluation, evaluation metrics used by the testing frameworks, driving model of AV under test and its input modality, and a complex multi-agent-based driving simulation environment. It is crucial to acknowledge these differences, as they are commonly encountered in the field of autonomous driving research~\cite{10064002}.

	\section{Related Work}~\label{sec:relatedwork}
	In this section, we summarize the literature for testing AVs and explain how our work differs from relevant related approaches. The main limitation of existing work is the lack of focus on first extracting the behavior representation of AV under test in order to find likely failure events without the need for expensive testing strategies. We present a learning algorithm in a black-box manner using inverse reinforcement learning on existing offline AV data.

	\subsection{Reward Modeling for Testing AVs}
	In general, IRL has played an important role in the advancement of autonomous system technologies. Its main application has been to model and replicate human driving behaviors~\cite{10160449,wu2020efficient}, as well as to plan behavior and motion~\cite{8968205,10073960}. Moreover, IRL has also been crucial in transferring knowledge to create new autonomous systems~\cite{mendez2018lifelong,tanwani2013transfer}.

	While IRL has been applied mainly to model behavior and transfer learning of expert policies, it has not yet been used for behavior representation in the testing of AVs. ReMAV's novel contribution is the use of IRL (especially AIRL~\cite{fu2017learning}) in behavior representation for testing and validating AVs, which to the best of our knowledge has not been performed before. We first utilize offline AV data to plan and attack noise perturbations purely based on the decision-making and behavior distribution on AV under test.

	\subsection{Multi-agent testing using perturbation attacks}
		
	Zhang et al.~\cite{DeepRoad} used GANs and metamorphic testing in DeepRoad to test autonomous driving systems, producing synthetic images for model robustness checks. Similarly, Deng et al.~\cite{9906558} used metamorphic testing in natural language processing to identify errors. DeepTest\cite{DeepTest} evaluates DNN-based AV models, generating real-world test cases, and applying transformations within the Udacity simulator. 	Haq et al.~\cite{9159088} presented a case study comparing the effectiveness of offline and online testing for deep learning-based AV models. Offline testing focuses on prediction errors against a dataset, while online testing looks for safety violations in driving scenarios. The study uses a pretrained driving model from the Udacity car simulator. As a limitation, the work needs an extension of multi-agent testing configurations within online and offline driving scenarios to observe which method will be more beneficial for multi-agent testing. Despite achieving great results, these methods are limited as they only test single-agent environments and overlook multi-agent testing.

	BayesOpt~\cite{R2} uses Bayesian optimization for autonomous vehicle testing, using Carla-based urban driving simulations by adding physical lines to create adversarial scenarios. Gangopadhyay et al.~\cite{8917103} also use Bayesian optimization to generate test cases for AVs. The method involves learning parameters by analyzing the system's output to create test scenarios that result in the AV failing. AutoFuzz~\cite{zhong2022neural} introduced a grammar-based fuzzing technique that uses the simulator API specification to generate semantically valid test scenarios for autonomous vehicle controllers. They compared their results with certain baselines, including AV-FUZZER~\cite{9251068} using four different driving scenarios. AV-FUZZER also uses the fuzzing technique by first incorporating scene descriptions and then applying a failure-coverage fuzzing algorithm to find scenarios that can violate different assertion statements. The proposed methods focus on generating semantically and temporally valid complex driving scenarios, but do not address all possible corner cases involving interactions with multiple independent agents.


	In a related study, reinforcement learning is applied to assess the robustness of AV in a simulated environment using adaptive stress testing (AST)~\cite{R34}. Another study by Koren et al.~\cite{9636072} extends the idea of RL stress testing by introducing a backward algorithm to find failure scenarios in a high-fidelity environment using expert demonstrations. This approach increases the search space and helps identify failure cases in AV driving policies. A different study by Wachi et al.~\cite{R8} used adversarial reinforcement learning to test a multi-agent driving simulation. This involves training multiple adversarial agents to compete against a single rule-based driving model. Although the results seem encouraging, the methods are only limited to rule-based driving systems within their experiments. DeepCollision~\cite{lu2022learning} as a recent work proposes a similar approach to AST and BayesOpt by learning how to configure their operating environment using reinforcement learning. They formalize it as an MDP and use deep Q-networks as the solution to test AI-based advanced AV systems. Compared to our work, we first utilize offline AV data to plan and attack minimal noise perturbations purely based on the decision-making on the AV under test.



	\subsection{Behavior analysis for reducing search space}
	Advsim~\cite{wang2021advsim} proposed a framework that generates worst-case scenarios for autonomous systems using an adversarial approach. This framework simulates plausible failure cases by perturbing sensor LiDAR data. Yulong et al.\cite{R32} utilized a GAN model to produce adversarial objects that can be used to attack LiDAR-based driving systems. Similarly, Delecki et al.~\cite{PALM} proposed a methodology to perform stress tests on LiDAR-based perception using a real-world driving dataset and various weather conditions to evaluate the performance of autonomous driving systems. 
	 Christian et al.~\cite{christian2023generating} addresses the limitations of existing methods for testing perception software, particularly LiDAR-based perception systems, by introducing a novel approach that generates realistic and diverse test cases through mutations while preserving realism invariants. As a limitation, they acknowledge the lack of AV behavior as part of the testing framework as approached in our work. LiRTest~\cite{guo2022lirtest} on the other hand introduced a LiDAR-based AV testing tool that addresses the limitations of conventional testing techniques. LiRTest implements AV-specific metamorphic relations and transformation operators to simulate various environmental factors and evaluate the performance of 3D object detection models under different driving conditions, effectively detecting erroneous behaviors and improving object detection precision. Ding et al.~\cite{9340696} proposed a framework for generating traffic scenarios by sampling from joint distributions of autoregressive building blocks. They develop an algorithm that uses the task algorithm to guide the generation module for creating safety-critical scenarios. 
	 	The authors perform sequential-based adversarial attacks which is time-consuming. The proposed ideas also do not take into account the computational cost of finding failure scenarios, especially in a complex multi-agent environment. We first investigate the existing state-action distribution of AV under test by building a reward model in an MDP manner before generating any sort of adversarial attack.
	

	Gambi et al.~\cite{Alessio} described using a technique called search-based testing to generate difficult virtual scenarios to test self-driving cars. These scenarios are used to examine the performance of AI driving models such as DeepDriving~\cite{chen2015deepdriving} and to systematically test lane-keeping systems. Song et al.~\cite{song2022critical} similarly proposed an end-to-end approach with a combination of various tools to identify critical states in two AV driving scenarios. DriveFuzz~\cite{drivefuzz} focused on holistic testing of autonomous driving systems by generating and mutating driving scenarios to uncover potential vulnerabilities, resulting in the discovery of new bugs in various layers of autonomous driving systems. DriveFuzz has successfully discovered new bugs in various layers of two autonomous driving systems. However, similar to most feedback-driven fuzzers that register a specific fitness function as feedback, DriveFuzz can have a local optima problem in the search space as a result of feedback guidance and ends up missing other potential bugs that are less related to the feedback. These methods do not incorporate reward modeling as a central part of their methodology. Also, these techniques may not specifically focus on narrowing down the search space for potential failures, potentially leading to a wide but less targeted testing scope. In contrast, ReMAV addresses AV as a multi-agent problem and adopts an MDP approach while learning the reward model using offline trajectories. This approach considers the complex dynamics of AVs and their interactions, enabling more effective decision-making in dynamic environments.

	\section{Conclusion}

	This paper presents a new testing framework for autonomous vehicles that uses offline data to analyze their behavior and set appropriate thresholds to detect the probability of failure events. The framework consists of three steps: first, it uses offline state-action pairs to build an abstract behavior model and identify states with uncertain driving decisions. Second, it employs a reward modeling technique to create a behavior representation that helps highlight regions of likely uncertain behavior, even when the standard autonomous vehicle performs well. Finally, the framework uses a disturbance model to test the vehicle's performance under minimal perturbation attacks, where driving decisions are less confident. Our framework finds an increase in failure events concerning vehicles, road objects, pedestrian collisions, and offroad steering incidents by the AVs under test. Compared to the two baselines, ReMAV is significantly more effective in generating these failure events in all metrics. An additional analysis with previous testing frameworks reveals that they need to improve the efficiency of training and testing, the total discovery of infractions, and the number of simulation steps to detect the first failure, compared to our method. In general, this framework provides a powerful tool for analyzing the behavior of autonomous vehicles and detecting potential failure events. Our study indicates that by using the proposed framework, it is possible to identify the vulnerabilities of autonomous vehicles and concentrate on attacking those areas, beginning with the basic perturbation models. We illustrate that our approach successfully detected failure events in the form of higher collision and steering errors compared to baselines.
	
	Essentially, ReMAV addresses the problem of expensive adversarial testing approaches that are performed without first analyzing AV's existing behavior. ReMAV also addresses the problem of finding and reducing the search space to only those states where AVs are less confident. Put simply, our testing framework proposes that once we can identify states with low reward values, we should only attack in those regions rather than attacking AVs from start to end of the simulation testing.
	
	 We believe that this testing framework can help in the fast adaptation of autonomous vehicles in the real world by providing a powerful tool to analyze their behavior and identify potential failure events. 
	
	\bibliographystyle{IEEEtran}
	\bibliography{sample-base}

	\vfill
	
\end{document}